\documentclass{article}

\usepackage[preprint]{neurips_2024}
\makeatletter
\renewcommand{\@noticestring}{}
\makeatother

\usepackage[utf8]{inputenc} 
\usepackage[T1]{fontenc}    
\usepackage{hyperref}       
\usepackage{url}            
\usepackage{booktabs}       
\usepackage{amsfonts}       
\usepackage{amsmath}        
\usepackage{nicefrac}       
\usepackage{microtype}      
\usepackage{xcolor}         
\usepackage{graphicx}       
\graphicspath{{figures/}}

\title{Kepler-Encoder-v0.1:\\
       Towards a Multimodal Embedding Model for Robots}

\author{%
  Ishneet Sukhvinder Singh
  \And
  Dhanoosh Pooranakumaran
  \And
  Alex Nguyen
  \And
  Jia Qi Yip\thanks{Correspondence: \texttt{jiaqi@menlo.ai}} \\
  Menlo Research \\
}

\begin{document}

\maketitle

\begin{abstract}
  A robot must understand the state of its own body, but a camera sees only part of it. Force and contact leave almost no trace in a single frame, and raw vision features read force at $R^2$ at or below $0.10$ on every robot we test. We present \textbf{Kepler-Encoder-v0.1}, a robot-first multimodal encoder that treats robot state as a modality and fuses vision, proprioception, and force/torque into a single shared latent with a learned-query cross-attention layer, trained self-supervised by masked cross-modal prediction under the LeJEPA/SIGReg objective~\citep{lejepa2025}. At evaluation only vision enters, which poses a sharp question. Does fusing state into training make the vision-only latent carry anything the pixels do not already contain? On the RH20T corpus~\citep{rh20t2023} the answer is yes, precisely where the camera is weakest. On held-out scenes, the vision-only latent recovers end-effector state, and force in particular, significantly above both raw frozen-ViT features and a compute-matched vision-only control on every sensored robot, though absolute force recovery at a single timestep is modest; on motor state, which the camera largely sees, it is statistically tied with the strongest vision baselines, and it is the only feature whose latent geometry tracks state. A single embodiment-agnostic encoder covers four robots, and a data-matched control shows this breadth reflects embodiment diversity rather than data volume. The frozen latent is directly useful. Its own cross-modal prediction error is a training-free invalid-state monitor (AUROC $0.90$ on out-of-range states, $0.69$ on scene-swapped states), and a diffusion decoder (PixNerd~\citep{pixnerd2026}) reconstructs the camera frame from the latent, confirming the spatial compression preserves world-state. This report validates the single-timestep case; native-rate temporal fusion is the next step.
\end{abstract}

\section{Introduction}
\label{sec:intro}

A robot that reasons about its body from video alone reasons about a body it cannot fully see. Force, contact, and joint torque leave little trace in a single RGB frame, and we confirm that force is nearly unreadable from vision alone (raw-feature $R^2$ at or below $0.10$ on every robot; Appendix~\ref{sec:stage1}, Table~\ref{tab:forceonly}). The common recipe (encode observations with a vision-only backbone and append robot state downstream as raw numbers~\citep{openvla2024}) inherits this blind spot. Yet on a robot, vision and state are not independent. On the internet an image is disembodied and could depict anything; on a robot, the images the camera produces are a narrow subset of all possible images, and which subset is fixed by joint configuration, gripper aperture, and contact. Vision and state are two views of one underlying body state, so encoding them jointly should let each inform the other.

We present \textbf{Kepler-Encoder-v0.1}, a robot-first multimodal encoder that maps vision and robot state (proprioception, force/torque, gripper) into a single shared latent. Because the two share a latent, the encoder partially recovers body state that lies outside the frame, and the representation carries force and proprioception rather than appearance alone. We train it without task labels by \emph{masked cross-modal prediction}, holding one modality out and predicting its representation from the rest, under the LeJEPA/SIGReg objective~\citep{lejepa2025}. Following JEPA~\citep{ijepa2023}, the target is a latent rather than pixels, which keeps the code focused on world-relevant structure instead of texture, the structure a robot needs to anticipate the consequences of its actions.

A robot's sensor set is variable in number, sampling rate, and composition across embodiments. We represent each reading as a token and compress the set with a single learned-query cross-attention block~\citep{perceiver2021}. Cost grows linearly in the number of tokens, and the output is a fixed-size latent regardless of how many sensors enter. One encoder therefore ingests a 6-DOF UR5 or a 7-DOF KUKA, with or without a force sensor, by masking absent inputs rather than maintaining a separate architecture per robot.

Our study is organized around one question, \emph{is vision enough?} That is, does a vision-only representation (or a compression of it) already encode the robot's physical state, so that fusing in force and proprioception buys nothing recoverable at test time? For the state the camera cannot see, it is not enough. There, the fused vision-only latent carries state that raw features and controlled baselines do not, while on camera-visible motor state it is statistically tied with the strongest vision baselines; one such encoder spans four robots; and the frozen latent is directly useful downstream, most notably as a training-free safety monitor. This report covers the encoder itself, the first step; native-rate temporal fusion (Appendix~\ref{sec:temporal}) and an action predictor built on the frozen latent are the natural next ones.

Our contributions are:
\begin{itemize}
  \item \textbf{A robot-first multimodal encoder.} We treat video, proprioception,
        and force/torque as first-class tokens, fuse them with a learned-query cross-attention layer, and train by masked cross-modal latent prediction under LeJEPA/SIGReg, with a robot-agnostic state representation that keeps every joint via masking rather than truncation (Section~\ref{sec:method}).
  \item \textbf{The vision-only latent carries body state the camera cannot see.}
        It reads end-effector state, and force in particular, out of pixels significantly better than raw frozen-ViT features on every sensored robot, and stays statistically tied with the strongest vision baselines on motor state, which the camera largely sees. Two controls, a compute-matched vision-only encoder and a PCA 256 compression of the pretrained ViT features, isolate the end-effector gain as cross-modal learning rather than in-domain training or dimensionality reduction, with force the clearest and most carefully bounded signal (Section~\ref{sec:mainresults}, Section~\ref{sec:ablations}).
  \item \textbf{One embodiment-agnostic encoder across four robots.} A single
        encoder matches per-robot specialists on their own robot and, for robots in the training mix, stays strong where specialists collapse; a data-budget-matched control shows the breadth reflects embodiment diversity rather than data volume, while on-diagonal parity partly reflects the larger training set (Section~\ref{sec:ablations}).
  \item \textbf{The frozen latent is directly useful downstream.} Its cross-modal
        prediction error is a training-free invalid-state / safety monitor (AUROC $0.90$ out-of-range, $0.69$ scene-swapped); and a latent-conditioned diffusion decoder (PixNerd) reconstructs the camera frame from the vision-only latent, confirming the spatial compression preserves world-state (Section~\ref{sec:downstream}).
\end{itemize}

\section{Related Work}
\label{sec:related}

\paragraph{Joint-embedding and predictive representations.}
Joint-embedding predictive architectures learn representations by predicting latent targets rather than reconstructing inputs, avoiding the pixel-memorization trap of autoencoders~\citep{ijepa2023,data2vec2022}. The paradigm has been applied to video~\citep{vjepa22025}, functional brain signals~\citep{jmaefmri2024}, multivariate time series~\citep{mtsjepa2026}, and robotic motor control~\citep{spatiotemporal2024}. Our training objective follows LeJEPA~\citep{lejepa2025}, whose SIGReg regularizer replaces the architectural anti-collapse heuristics (stop-gradient, predictor asymmetry, teacher momentum) with a distributional constraint. On the multimodal side, masked cross-modal prediction appears in M3-JEPA~\citep{m3jepa2024}, 4M~\citep{fourm2023}, MultiMAE~\citep{multimae2022}, and data2vec~\citep{data2vec2022}; we adopt the predict-one-modality-from-the-rest structure but in \emph{latent} space with SIGReg, over robot sensor streams rather than images/text.

\paragraph{Robot multi-sensor representation.}
Vision-language-action models~\citep{openvla2024} produce a vision-centric representation and treat proprioception as an appended input; heterogeneous pre-trained transformers align proprioceptive-visual streams across embodiments~\citep{hpt2024}. A separate line learns tactile and multisensory touch representations~\citep{sparsh2024,sparshx2025,seehearfeel2022,m3l2023}. The closest work, MSDP~\citep{msdp2025}, pre-trains a multi-sensory dynamics model but \emph{assumes synchronized observations}, precisely the multi-rate assumption our tokenization is designed to remove in the native-rate extension (Appendix~\ref{sec:temporal}). Action tokenization for sequence models~\citep{vqvla2025,fast2025,actioncodec2026,mojito2025} is complementary, compressing actions for policies, whereas we compress \emph{observations} into a reusable state latent. Beyond robotics, a wearable-health foundation model~\citep{sensorfm2026} fuses heterogeneous physiological streams at population scale, evidence that heterogeneous-sensor pretraining generalizes beyond robots.

\paragraph{Latent-query cross-attention.}
Our fuser summarizes a variable-size token set with a fixed number of learned queries, compressing $N$ tokens into $M$ latents at $O(NM)$ rather than the $O(N^2)$ of self-attention. The pattern originates with the Set Transformer's induced set-attention block~\citep{settransformer2019} and was generalized by the Perceiver family~\citep{perceiver2021,perceiverio2021,perceiverar2022}. The same learned-query mechanism recurs across detection and segmentation~\citep{detr2020,mask2former2022}, resamplers that condense visual features for frozen language models~\citep{flamingo2022,blip2_2023}, audio-visual fusion~\citep{mbt2021}, and speaker verification~\citep{acanet2023}.

\paragraph{Continuous-time and irregular multimodal fusion.}
The continuous-time embedding of each token's timestamp borrows from the irregular time-series literature, including mTAN~\citep{mtan2021}, Time2Vec~\citep{time2vec2019}, SeFT~\citep{seft2019}, and ContiFormer~\citep{contiformer2024}. COPER couples a continuous-time embedding with a Perceiver~\citep{coper2022}, and FuseMoE~\citep{fusemoe2024} and UTDE~\citep{utde2022} fuse irregular multimodal streams, but all in the electronic-health-record domain. The planned extension (Appendix~\ref{sec:temporal}) transplants the continuous-time-token idea into robot sensor fusion, where the rate mismatch (100~Hz force vs.\ 10~Hz vision) is the defining difficulty.

\paragraph{Image and video autoencoders; compression.}
Compression-oriented and diffusion-based video autoencoders~\citep{adaptive1dvdae2026,rvmae2026,videoae2601,videoae2604, multimodalcompression2022} and neural codecs repurposed as tokenizers~\citep{biosignaltokenizers2025} inform the (optional) decoder and the compression framing, but not the training signal, since reconstruction is deliberately kept out of the objective.

\paragraph{Robot interaction datasets.}
We train on RH20T~\citep{rh20t2023}, and situate our data choices relative to humanoid~\citep{humanoidseveryday2025}, large-scale manipulation~\citep{agibotworld2025}, mobile manipulation~\citep{airoamoma2025}, and contact-rich force-aware collections~\citep{forcevla22026}.

\paragraph{Downstream validation target.}
Auxiliary latent-prediction objectives improve VLA policies when the target encoder is strong. FLARE~\citep{flare2025} predicts the latent of a future observation and shows the target encoder's quality is the deciding factor. This is a natural downstream slot for a multimodal encoder such as ours, and for the native-rate extension it is built toward.

\section{Method}
\label{sec:method}

\subsection{Problem formulation}
\label{sec:problem}
At time $t$ a robot emits a set of heterogeneous sensor streams, namely an RGB frame $x_v$, proprioceptive state $x_m$ (joint angles, velocities, gripper aperture), and end-effector signals $x_e$ (force/torque, TCP pose). These streams are partial views of one underlying body state. We seek an encoder $f_\theta$ that fuses the available streams into a single latent $z$ (a unified representation of the robot and its perceivable environment) and, at evaluation, recovers that latent from vision alone:
\begin{equation}
  \label{eq:encoder}
  z = f_\theta(x_v, x_m, x_e), \qquad z_v = f_\theta(x_v, \emptyset, \emptyset),
\end{equation}
where $x_v, x_m, x_e$ are the vision, motor, and end-effector streams and $\emptyset$ marks a \emph{dropped} stream, one for which no input is provided at evaluation. Because $z_v$ is the exported evaluation embedding, the representation must be genuinely cross-modal; we adapt the spatial masking of I-JEPA~\citep{ijepa2023} to the modality dimension. Writing $z^{(\backslash k)}$ for the fused latent with modality $k$ held out, and $\bar f_{\bar\theta}$ for an exponential-moving-average target copy of the encoder, the learning objective is
\begin{equation}
  \label{eq:objective}
  \mathcal{L}_{\text{pred}}
  = \sum_{k \in \{v,\,m,\,e\}}
    \big\lVert\, h_k\!\big(z^{(\backslash k)}\big) - \operatorname{sg}\!\big[\bar f_{\bar\theta}(x_k)\big] \,\big\rVert_2^2,
\end{equation}
with $\operatorname{sg}[\cdot]$ the stop-gradient and $h_k$ a per-modality predictor head that maps the fused latent back to modality $k$'s target embedding. The full training loss adds anti-collapse regularization to \eqref{eq:objective}, detailed in Section~\ref{sec:objective}. The encoder must therefore (i) retain the force and proprioceptive structure the camera cannot reveal, (ii) keep $z_v$ informative of body state despite dropping $x_m, x_e$ at test time, and (iii) share one architecture across embodiments, tolerating hardware-absent sensors by masking.

Which images the camera produces is determined by joint configuration, gripper aperture, and contact, and force leaves essentially no trace in a single frame ($R^2 \approx 0$; Appendix~\ref{sec:stage1}). Crucially, this force is a \emph{directly sensed} instantaneous state variable, read from the wrist force/torque sensor at each tick rather than differentiated from motion, so a single frame is a well-posed target for it, exactly as for joint angle or gripper aperture. Velocity and acceleration are the opposite case, genuine time-derivatives of position that a single frame cannot determine and that lie outside the single-timestep model of this report (Appendix~\ref{sec:temporal}). The stream count, native sampling rates, and the sensor suite itself vary across embodiments, so no fixed state vector spans the corpus. Appendix~\ref{sec:notation} collects every symbol used in this section.

\subsection{Token representation}
\label{sec:tokenization}
Every sensor reading is embedded as a token that carries its tokenized value, a learned modality embedding, a spatial position, and a continuous-time embedding of its timestamp (constant in the single-timestep model of this report; Appendix~\ref{sec:temporal}).

\begin{itemize}
  \item \textbf{Vision.} Patch tokens from a \emph{frozen} LeJEPA ViT-B/16
        (196 patches $\times$ 768-d); we write $F_v$ for these patch features of the frame $x_v$. The backbone is frozen on the evidence of Appendix~\ref{sec:stage1} (finetuning on robot video degrades the encoder and adds nothing robot-relevant), not by assumption. Only the fusion head trains ($\sim$2M parameters).
  \item \textbf{Proprioception (motor).} A robot-agnostic $8\times3$ grid, where rows
        0--6 are joints and row 7 is the gripper. Channels use scale-respecting encodings, namely $[\sin q, \cos q, \mathrm{symlog}\,\dot q]$ for joints and symlog width for the gripper. Angles use $\sin/\cos$ to remove the $2\pi$ wrap, unbounded quantities use symlog, and rotations use the continuous 6-D representation~\citep{rot6d2019}.
  \item \textbf{End-effector (ee).} A $13\times15$ block of native-rate samples
        in the interval $[\text{tick}_k, \text{tick}_{k+1})$, comprising symlog force/torque (6), symlog TCP translation (3), and TCP 6-D rotation (6), zero-padded to a fixed window.
\end{itemize}

\paragraph{Robot-agnostic state via masking rather than truncation.}
Different robots expose different joint counts (6 for UR5, 7 for Flexiv, Franka, and KUKA) and different sensor suites (Franka has no physical force/torque sensor). Rather than define a per-robot state vector, we fix the $8\times3$ motor grid and the $13\times15$ ee block for \emph{all} robots and mask the rows and channels a given robot does not populate. This choice has three consequences. One architecture spans every embodiment; joints remain first-class tokens rather than being truncated away; and a hardware-absent sensor enters through the same validity masks as any other invalid reading, zero-padded values with validity bits \eqref{eq:motorfeat}. That data-level masking is a different operation from the attention-level hiding the training objective uses (Section~\ref{sec:objective}), but it removes the stream from fusion with the same effect, so Franka's hardware-absent force sensor provides a built-in robustness test rather than a special case.

\subsection{Architecture: the cross-attention fuser}
\label{sec:arch}

\begin{figure}[!htbp]
  \centering
  \includegraphics[width=\linewidth]{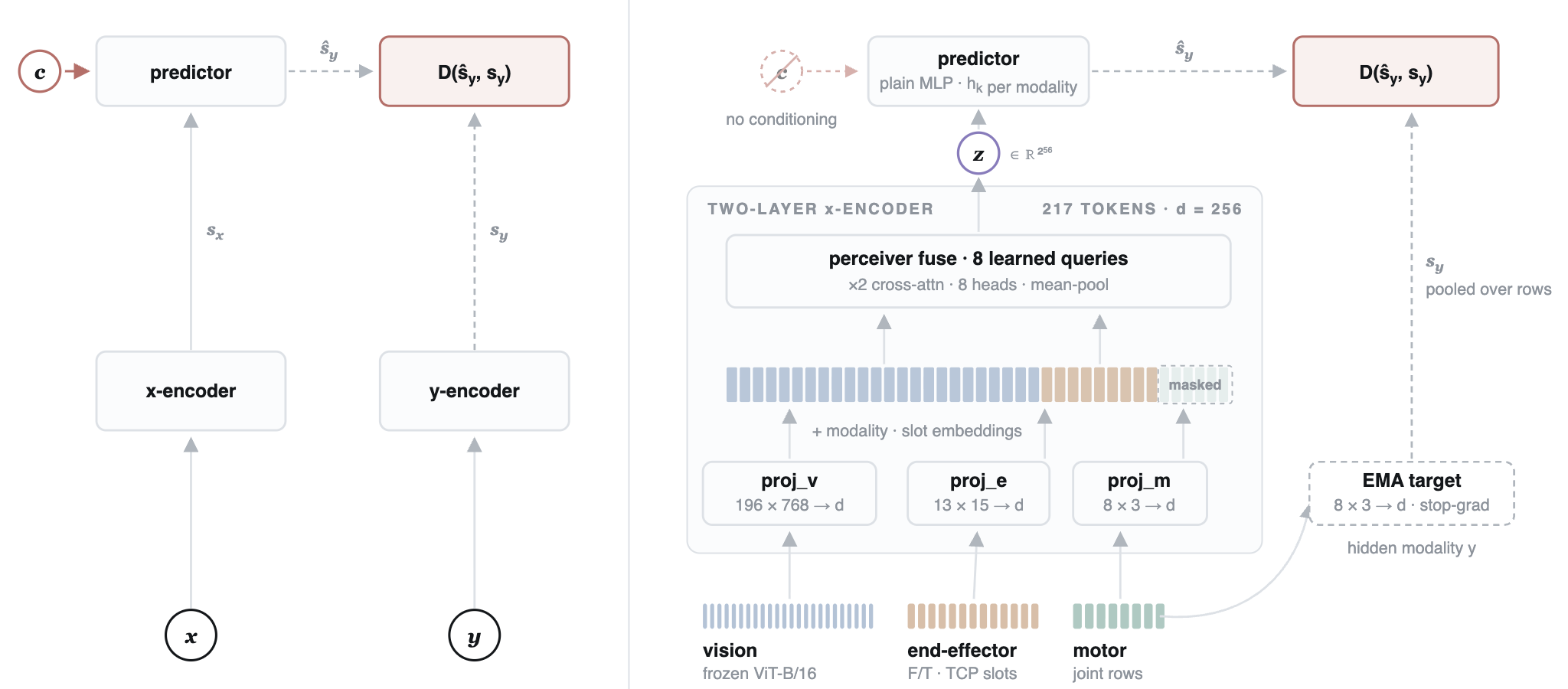}
  \caption{The training objective as a joint-embedding predictive architecture. \emph{Left:} the generic JEPA template, in which an x-encoder and a y-encoder embed two views and a predictor maps one embedding to the other under a distance $D(\hat s_y, s_y)$~\citep{ijepa2023}. \emph{Right:} our instantiation. The visible modalities are projected (\emph{proj\_v}, \emph{proj\_e}, \emph{proj\_m}), tagged with modality and slot embeddings, and fused by the two-block Perceiver over 217 tokens into $z \in \mathbb{R}^{256}$; a per-modality MLP head $h_k$ then predicts the held-out modality's target embedding, produced by an EMA copy of that modality's projection under stop-gradient (motor shown hidden; the held-out modality rotates over vision, motor, and ee). The template's conditioning variable $c$ is unused. The loss is $\mathrm{MSE}(\text{prediction},\text{target})$ plus per-modality and joint SIGReg (LeJEPA), which prevents collapse.}
  \label{fig:loss}
\end{figure}

\paragraph{Forward pass.} The fuser is a single asymmetric cross-attention
block in which a fixed set of learned queries reads from the token set (Section~\ref{sec:related}), and the right panel of Figure~\ref{fig:loss} traces it end to end. The three projected streams are concatenated into a single context set of $N = N_v + N_m + N_e$ tokens ($N_v$ vision, $N_m$ motor, and $N_e$ ee), heterogeneous in origin but homogeneous in width $d$. A set of $M$ learned query vectors cross-attends over this set, and in each of $L$ depth blocks the queries attend to the context and pass through an FFN (pre-norm residuals), after which they are mean-pooled into a single fixed-size latent $z\in\mathbb{R}^{d}$. The per-stream token counts, the latent width $d$, the query count $M$, and the depth $L$ are hyperparameters, set to the values reported in the experiments (Section~\ref{sec:experiments}). Attention masks serve two roles, encoding padding and validity as well as the modality masking of the training objective. The two mask types use opposite sign conventions, since chunk masks mark valid entries as \texttt{True}, whereas cross-attention masks mark blocked entries as \texttt{True}, so one is inverted before use.

\paragraph{Formal specification.} The fuser realizes the encoder $f_\theta$ of \eqref{eq:encoder} after tokenization. Abusing notation, $f_\theta(x_v, x_m, x_e) = \mathrm{Fuse}(C, \mathcal{M}_{\mathrm{valid}})$ when every stream is present, and a stream dropped at evaluation corresponds to hiding its tokens, as in \eqref{eq:latents} below. The motor stream first zeroes the channels a robot does not populate and appends the validity bits as extra features, so the projection sees both the value and the fact that it is present,
\begin{equation}
  \label{eq:motorfeat}
  \phi_m(x_m, \mathbf{v}_m) = \big[\, x_m \odot \mathbf{v}_m \;\Vert\; \mathbf{v}_m \,\big],
\end{equation}
where $\mathbf{v}_m$ is the motor validity mask, matching $x_m$ in shape, and $\Vert$ is channel concatenation. Each stream is then linearly projected to the shared width $d$ and tagged with a learned modality embedding $\mathbf{e}_{(\cdot)}$ and, for the structured streams, a learned position embedding $\mathbf{p}_{(\cdot)}$,
\begin{align}
  \label{eq:proj}
  T_v &= W_v\, F_v + \mathbf{e}_v, &
  T_m &= W_m\, \phi_m(x_m,\mathbf{v}_m) + \mathbf{e}_m + \mathbf{p}_m, &
  T_e &= W_e\, x_e + \mathbf{e}_e + \mathbf{p}_e,
\end{align}
producing $N_v$, $N_m$, and $N_e$ tokens respectively, each of width $d$. The context set is their concatenation,
\begin{equation}
  \label{eq:context}
  C = \big[\, T_v \;\Vert\; T_m \;\Vert\; T_e \,\big] \in \mathbb{R}^{N\times d}, \qquad N = N_v + N_m + N_e.
\end{equation}
The fuser $\mathrm{Fuse}(C, \mathcal{M})$ carries $M$ learned queries $Q^{(0)}=\{\xi_i\}_{i=1}^{M}$ through $L$ pre-norm blocks that cross-attend to $C$ under a boolean block-mask $\mathcal{M}$ (entries set to $-\infty$ in the attention logits),
\begin{align}
  \label{eq:fuse}
  \tilde Q^{(\ell)} &= Q^{(\ell-1)} + \mathrm{CrossAttn}\big(\mathrm{LN}(Q^{(\ell-1)}),\, C,\, \mathcal{M}\big), &
  Q^{(\ell)} &= \tilde Q^{(\ell)} + \mathrm{FFN}\big(\mathrm{LN}(\tilde Q^{(\ell)})\big),
\end{align}
for $\ell = 1,\dots,L$, and the fused latent is the mean over queries,
\begin{equation}
  \label{eq:pool}
  \mathrm{Fuse}(C,\mathcal{M}) = \frac{1}{M}\sum_{i=1}^{M} Q^{(L)}_i \in \mathbb{R}^{d}.
\end{equation}
Writing $\mathcal{M}_{\mathrm{valid}}$ for the mask that blocks only invalid and padding tokens, and $\mathcal{M}_{\text{hide}(S)}$ for the mask that additionally blocks the tokens of every modality in $S$, the training-time fused latent uses the full context and the held-out-modality latents hide one stream,
\begin{equation}
  \label{eq:latents}
  z = \mathrm{Fuse}\big(C, \mathcal{M}_{\mathrm{valid}}\big), \qquad
  z^{(\backslash k)} = \mathrm{Fuse}\big(C, \mathcal{M}_{\text{hide}(\{k\})}\big), \qquad
  z_v = \mathrm{Fuse}\big(C, \mathcal{M}_{\text{hide}(\{m,e\})}\big),
\end{equation}
where $z_v$, the vision-only latent obtained by hiding motor and ee, is the exported evaluation embedding. The held-out latents $z^{(\backslash k)}$ feed the predictor heads of \eqref{eq:objective}.

\begin{figure}[!htbp]
  \centering
  \includegraphics[width=\linewidth]{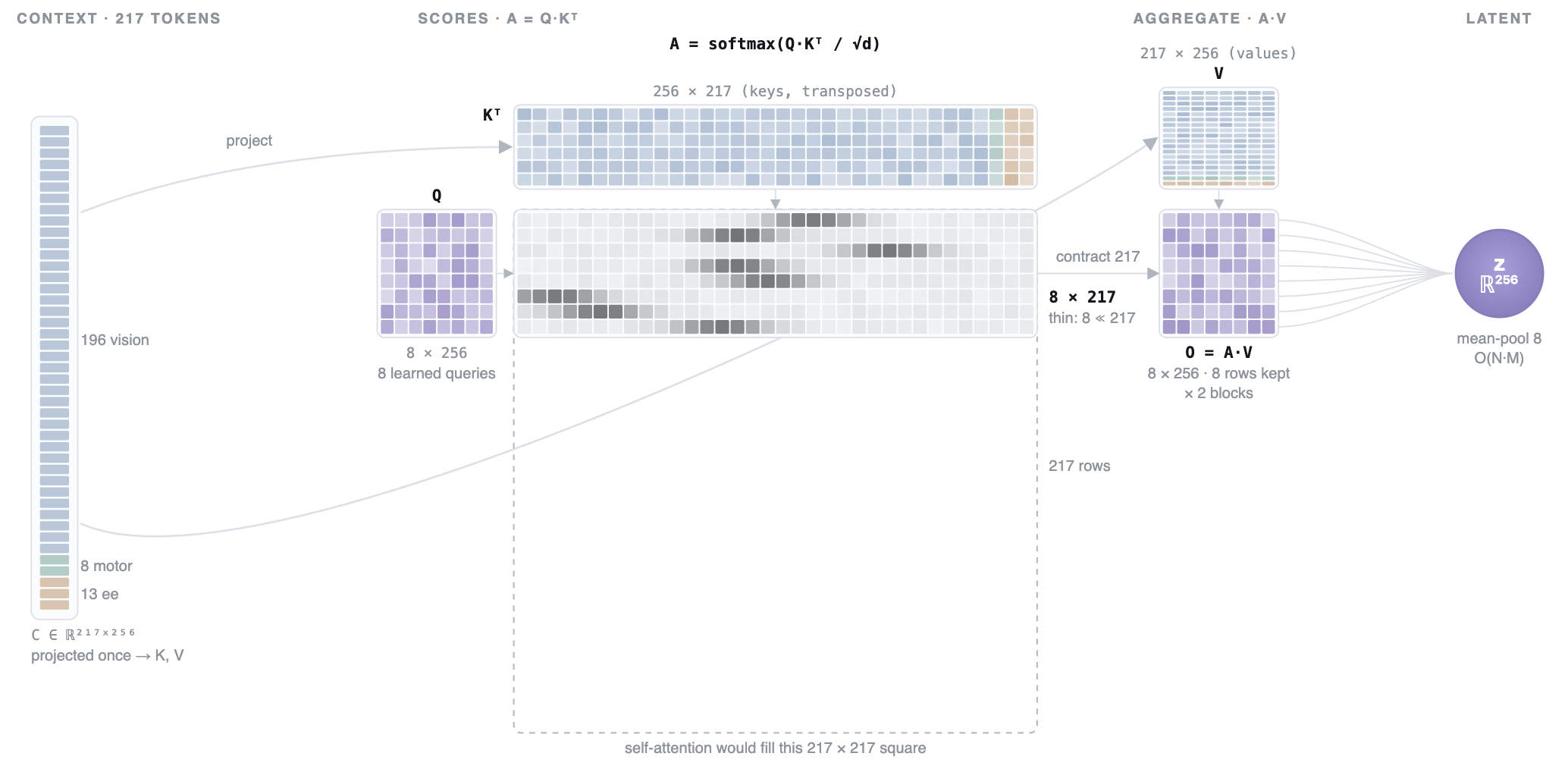}
  \caption{The cross-attention bottleneck at matrix level, with the sizes of the reported instantiation. The context $C \in \mathbb{R}^{217\times256}$ (196 vision, 8 motor, 13 ee tokens) is projected once to keys and values; the $M=8$ learned queries score it as $A = \mathrm{softmax}(QK^\top/\sqrt{d})$, an $8\times217$ matrix rather than the $217\times217$ square that self-attention would fill (dashed outline), and the output $O = AV$ keeps 8 rows per block before the mean-pool to $z \in \mathbb{R}^{256}$. Cost therefore grows as $O(NM)$ with $M \ll N$.}
  \label{fig:crossattn}
\end{figure}

\paragraph{Why a latent-query bottleneck.} Three properties suit this block to heterogeneous robot streams, over a full self-attention transformer or late concatenation of per-modality encoders. \emph{(i)~Compute decoupled from input size:} cross-attention from $M$ queries to $N$ tokens costs $O(NM)$ (Figure~\ref{fig:crossattn}), so adding a sensor or more patches grows the context linearly and leaves the latent width unchanged. \emph{(ii)~Fixed-size latent under variable input:} $z$ has the same shape whether a robot exposes 6 or 7 joints, carries a force sensor or not, or (at evaluation) sees vision alone, so masking rather than per-robot truncation carries embodiment-agnosticism; a hardware-absent sensor amounts to blocked columns, avoiding the per-embodiment ``stems'' HPT needs~\citep{hpt2024}. \emph{(iii)~Set-structured fusion:} position and modality live in the token embeddings rather than a fixed layout, so one block ingests any token set, and every query reads all modalities at once~\citep{perceiver2021}.

\subsection{Training objective}
\label{sec:objective}
Training combines a cross-modal predictive signal with two anti-collapse regularizers. Let $z$ denote the fused latent and $z^{(\backslash k)}$ the fused latent when modality $k$ is held out, following \eqref{eq:latents}.

\begin{enumerate}
  \item \textbf{Masked cross-modal latent prediction (the learning signal).}
        Hold out one modality (vision, motor, or ee) and predict its exponential-moving-average \emph{target} latent from the other two. The prediction is made by the per-modality head $h_k$ from the held-out-modality latent $z^{(\backslash k)}$, and its target is the frozen EMA encoder's embedding $\operatorname{sg}[\bar f_{\bar\theta}(x_k)]$, not the raw input $x_k$. This applies JEPA's predict-do-not-equate principle~\citep{ijepa2023,data2vec2022} across modalities, in the manner of masked multimodal prediction~\citep{m3jepa2024,fourm2023,multimae2022} but carried out in \emph{latent} space. It forces each stream to encode structure relevant to the others (vision learns force-relevant cues because it is trained to predict force), avoids intersection collapse, and confers robustness to dropped or hardware-absent streams by construction. A shared encoder without this signal underperforms single-modality baselines, so the cross-modal loss is necessary rather than optional. The ee prediction term is evaluated only over samples whose robot carries a force sensor, since embodiments without one (for example Franka) supply no ee target to predict.
  \item \textbf{Per-modality SIGReg.} A SIGReg (SlicedEppsPulley) penalty applied
        to each modality's embeddings before fusion. It prevents per-modality collapse and acts as a magnitude standardizer, making modalities commensurate before they are fused, a standardization that naive concatenation of separately-trained encoders omits. As with the prediction term, the ee penalty is applied only when a batch holds enough force-sensored samples, so vision and motor are always regularized while ee is regularized conditionally.
  \item \textbf{Joint SIGReg.} The same penalty applied to the fused latent $z$,
        with the aim of keeping it high-rank and expressive rather than collapsed. At the current scale its empirical effect is small (Appendix~\ref{sec:moreabl}); we retain it as insurance against collapse at wider bottlenecks and larger data, where anti-collapse pressure on $z$ matters more.
\end{enumerate}

The EMA target encoder is a slow copy of the online encoder, the momentum-target mechanism that stabilizes predictive SSL~\citep{data2vec2022}; SIGReg is the \texttt{SlicedEppsPulley} objective from LeJEPA~\citep{lejepa2025}, the explicit distributional anti-collapse constraint described in Section~\ref{sec:related}, and collapse is monitored directly with RankMe~\citep{rankme2023}. Only the fusion head and the per-modality projections train; the vision backbone stays frozen.

A decoder is deliberately excluded from this objective, because training on reconstruction would pull the latent back toward pixel memorization, the autoencoder failure mode this design is intended to avoid. Decoders reappear only at evaluation, as probes of what the frozen latent stores (Section~\ref{sec:storage}, Section~\ref{sec:qualitative}).

\section{Data}
\label{sec:data}

\paragraph{Corpus.}
We train on RH20T~\citep{rh20t2023}, a large real-robot manipulation dataset with 7 configurations spanning 4 embodiments with different degrees of freedom, grippers, and sensor suites (Table~\ref{tab:cfgs}). Each scene provides multi-view RGB with per-frame timestamps and robot state (joint position/velocity, TCP pose, force/torque, gripper). The sharded corpus is $\sim$54.3M frames ($\sim$4~TB) across 12{,}776 scenes.

\begin{table}[!htbp]
  \centering
  \caption{The 7 RH20T configurations are 4 robots with heterogeneous sensor
  suites. ``Joint vec'' is the length of \texttt{joint.npy}, which concatenates position with velocity (and torque on KUKA), so joint \emph{positions} are always the first DOF entries. Franka (cfg5) has no physical force/torque sensor; its end-effector channels are masked, serving as a built-in test of hardware-absent sensors.}
  \label{tab:cfgs}
  \begin{tabular}{clccccr}
    \toprule
    cfg & robot & DOF & gripper & F/T sensor & joint vec & robot scenes \\
    \midrule
    1 & Flexiv Rizon & 7 & Dahuan AG-95 & dahuan & 14 & 4{,}268 \\
    2 & Flexiv Rizon & 7 & Dahuan AG-95 & dahuan & 14 & 1{,}792 \\
    3 & UR5          & 6 & WSG-50       & ati    & 6  & 799   \\
    4 & UR5          & 6 & Robotiq 2F-85& ati    & 6  & 2{,}194 \\
    5 & Franka       & 7 & Franka       & none   & 14 & 1{,}321 \\
    6 & KUKA iiwa    & 7 & Robotiq 2F-85& ati    & 21 & 1{,}485 \\
    7 & KUKA iiwa    & 7 & Robotiq 2F-85& ati    & 21 & 917   \\
    \midrule
    \multicolumn{6}{r}{Total robot scenes} & 12{,}776 \\
    \bottomrule
  \end{tabular}
\end{table}

\paragraph{The sensor streams in plain terms.}
Each scene delivers four kinds of measurement, and the rest of the paper leans on all of them, so we spell out what each one physically is.
\begin{itemize}
  \item \textbf{RGB frames.} What the external camera sees, one timestamped color image per tick. This is the only stream available at evaluation.
  \item \textbf{Joint state.} A robot arm is a chain of motorized rotary joints. The joint angle $q_i$ says how far joint $i$ is currently rotated, and the joint velocity $\dot q_i$ says how fast it is rotating; together the angles fix the arm's shape in space. The gripper aperture is the distance between the gripper's fingers, from fully open to fully closed.
  \item \textbf{TCP pose.} The tool center point (TCP) is the working reference point of the end-effector, conventionally placed between the gripper fingertips. Its pose is that point's 3-D position plus its 3-D orientation in the workspace, in plain terms where the hand is and which way it points. Unlike joint state, which describes the arm's internal configuration, the TCP pose describes the hand's placement in the world.
  \item \textbf{Force/torque.} A sensor in the wrist reports six numbers, the force pushing along each of three axes and the torque twisting about each of them. This is the robot's sense of touch and effort at the wrist, the signal that distinguishes pressing on a surface from hovering above it. Franka (cfg5) has no such sensor.
\end{itemize}

\paragraph{Timing.}
The shipped state streams are camera-aligned (their timestamps match the color frames), so the released data runs at an effective $\sim$7--15~Hz (varying by config and scene) rather than the raw sensors' native rate. A separate high-frequency stream (wrist force/torque and TCP) runs at 100--125~Hz, absent for cfg5 and empty for a third of cfg3. This irregular, multi-rate timing is why the model cannot assume a fixed clock. We split scenes at recording gaps $>500$~ms before chunking.

\paragraph{Chunk packet.}
We convert scenes into tick-anchored chunks (native samples only, no interpolation), each a robot-agnostic packet (Table~\ref{tab:packet}): frozen ViT patch tokens, a masked $8\times3$ motor grid, a masked $13\times15$ end-effector block, a robot id, and a tick timestamp (cached, so the temporal extension needs no re-preprocessing). Chunks are precomputed to per-config caches.

\begin{table}[!htbp]
  \centering
  \caption{The chunk packet emitted by the dataloader (per sample; $B$ is batch
  size). Masks mark \emph{valid} entries; the singleton time axis (index 1) is a placeholder for the continuous-time extension discussed in Appendix~\ref{sec:temporal}. The $\sim$12.4\% of the corpus with an all-False \texttt{ee\_mask} is all of cfg5 plus the hf-empty third of cfg3.}
  \label{tab:packet}
  \small
  \begin{tabular}{lll}
    \toprule
    key & shape & contents \\
    \midrule
    \texttt{rgb}       & $[B,1,196,768]$ & frozen ViT-B/16 patch tokens of the chunk's frame \\
    \texttt{motor}     & $[B,1,8,3]$     & 8 actuator rows (joints 1--7, gripper) $\times$ $[\sin q,\cos q,\mathrm{symlog}\,\dot q]$ \\
    \texttt{motor\_mask}& $[B,8,3]$      & valid rows/channels (UR5: seventh joint row and velocities masked) \\
    \texttt{ee}        & $[B,13,15]$     & 13 wrist samples/tick $\times$ symlog F/T (6), TCP pos.\ (3), 6-D rot.\ (6) \\
    \texttt{ee\_mask}  & $[B,13]$        & valid sample rows; all-False for cfg5 and hf-empty cfg3 ($\sim$12.4\%) \\
    \texttt{robot\_id} & $[B]$           & 0 flexiv, 1 ur5, 2 franka, 3 kuka \\
    \texttt{ts}        & $[B]$           & tick timestamp (ms), for $\Delta t$-based selection \\
    \bottomrule
  \end{tabular}
\end{table}

\paragraph{Reading the encodings.}
The packet stores transforms of the raw values, not the values themselves. The $8\times3$ motor grid has one row per actuator, with joints in rows 0--6 (the seventh joint row masked on 6-DOF robots) and the gripper in row 7. Its 3 columns are \emph{not} three physical quantities. A joint row holds $[\sin q, \cos q, \mathrm{symlog}\,\dot q]$, the angle in the first two columns through its encoding and the velocity in the third. The gripper row carries only its symlog width, in the first column. Joint torque, which the KUKA vector ships, is deliberately excluded. The $13\times15$ ee block holds the wrist stream instead, one row per native-rate sample landing between the current camera tick and the next (zero-padded, with \texttt{ee\_mask} marking the filled rows). Each ee row's 15 features are the six symlog force/torque readings, the three symlog TCP position coordinates, and the TCP orientation's six rotation numbers. Each transform fixes a specific numerical pathology of its quantity, as follows.
\begin{itemize}
  \item \textbf{Joint angles $\rightarrow [\sin q, \cos q]$.} Here $q$ is one joint's rotation angle in radians, read from the shipped joint-position vector (\texttt{joint.npy}), and every joint of the arm contributes its own pair to the motor grid. The raw angle wraps around, so $359^\circ$ and $1^\circ$ are physical neighbors yet numerically far apart, and a regression target with that jump is needlessly hard. Storing the pair $(\sin q, \cos q)$, the coordinates of the tip of a clock hand pointing at angle $q$, places wrapped neighbors close together and removes the discontinuity. The raw angle is recovered exactly as $q = \mathrm{atan2}(\sin q, \cos q)$, which is how the decoded joint angles of Figure~\ref{fig:statedecode} are produced.
  \item \textbf{Joint velocities, gripper width, force/torque, TCP position $\rightarrow \mathrm{symlog}$.} These raw values (the joint's angular velocity $\dot q$ where the robot reports one, the commanded finger-opening width from \texttt{gripper.npy}, the six wrist force and torque readings, and the TCP's position coordinates from \texttt{tcp\_base.npy}) share a different pathology. They are unbounded and heavy-tailed, mostly small values with occasional large spikes, for example at contact. Each is stored as $\mathrm{symlog}(x) = \mathrm{sign}(x)\,\log(1+|x|)$, a log scale that also accepts negative values; near zero it is almost the identity, so small readings stay distinguishable, while spikes are compressed so they do not dominate the loss. The raw value is recovered as $\mathrm{sign}(y)(e^{|y|}-1)$. One caveat carries over from the hardware, since the three gripper models report widths on their own scales, raw widths are not comparable across configs.
  \item \textbf{TCP orientation $\rightarrow$ 6-D rotation.} The corpus ships the TCP orientation as a quaternion (the last four entries of the 7-D pose in \texttt{tcp\_base.npy}). A 3-D rotation suffers the same wrap-around problem as a single angle, in three dimensions at once, and quaternions additionally represent every rotation twice. We therefore convert to the first two columns of the rotation matrix, six numbers that vary continuously with the physical orientation~\citep{rot6d2019}.
\end{itemize}

\paragraph{Splits and filtering.}
One user repeating a task yields $\sim$10 near-duplicate scenes, so we hold out by \textbf{(config, task, user) group} (stratified per config, $\sim$30\% held out, frozen in a committed CSV) and evaluate only on held-out groups, so no near-copy leaks into training. We use one deterministic external camera per scene (the wrist camera is excluded; multi-view is future work, framed as cross-view prediction). Two documented traps are filtered out, namely the $543$ \texttt{\_human\_2} scenes, which carry no robot state, and the $57$ scenes ($\sim$0.4\%) missing \texttt{joint.npy}.

\section{Experiments}
\label{sec:experiments}
Our experiments test a single hypothesis. The \emph{vision-sufficiency} null asserts that a vision-only representation, or a low-dimensional compression of one, already captures the robot's physical state, so that fusing proprioception and force during training adds nothing recoverable from vision at test time. Force makes the null non-trivial because it leaves almost no trace in a single frame (raw-feature $R^2$ at or below $0.10$ on every robot; Table~\ref{tab:forceonly}). We reject it in three steps. Section~\ref{sec:setup} presents the headline comparison, scoring the encoder against the vision-only baselines of Section~\ref{sec:baselines} with every result averaged across embodiments. Section~\ref{sec:ablations} breaks the comparison out by embodiment and isolates each ingredient of the result, covering the cross-embodiment transfer matrix, a compute-matched vision-only control, the separation of force from pose, and a data-budget-matched control. Section~\ref{sec:downstream} then shows that the frozen encoder is directly useful downstream with no retraining.

\subsection{Training results}
\label{sec:setup}
Unless noted otherwise, every result in this section concerns \textbf{Kepler-Encoder-v0.1}, a single encoder trained on all seven RH20T configurations (Section~\ref{sec:data}), abbreviated \textbf{KEv0.1} in tables. The instantiation sets $d=256$, $M=8$ queries, and $L=2$ cross-attention blocks (8 heads) over $N_v=196$, $N_m=8$, and $N_e=13$ tokens, and trains for 40 epochs at batch 256 with AdamW (learning rate $10^{-3}$, weight decay $10^{-4}$), repeated over 5 seeds. Only the fusion head and the per-modality projections train ($\sim$2M parameters); the ViT backbone stays frozen. Evaluation uses the committed group-held-out splits (Section~\ref{sec:data}), probes are fit on train rows and scored on held-out rows only, and error bars are $\pm$std over seeds. Per-robot specialist encoders, trained identically on single embodiments, appear in the ablation study (Section~\ref{sec:ablations}).

\subsubsection{Training loss}
\begin{figure}[!htbp]
  \centering
  \includegraphics[width=0.72\linewidth]{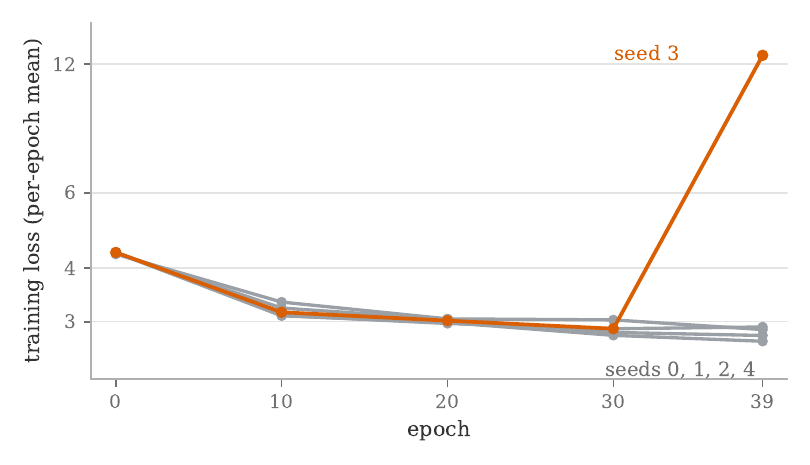}
  \caption{Training loss of Kepler-Encoder-v0.1 (per-epoch mean of the total
  objective, prediction $+$ SIGReg; logged every ten epochs; log scale). Four seeds descend from $\approx$4.3 and settle at 2.7--2.9. Seed 3 tracks them through epoch 30 and spikes to $12.57$ in the final epochs, yet its linear-probe accuracy is unchanged, illustrating why encoder quality is not read off this curve.}
  \label{fig:trainloss}
\end{figure}

The training loss sums the masked cross-modal prediction term and the SIGReg penalties (Section~\ref{sec:objective}). Figure~\ref{fig:trainloss} shows its per-epoch mean for the five seeds of Kepler-Encoder-v0.1, logged every ten epochs. Optimization is stable and consistent across seeds, with most of the decrease arriving in the first ten epochs ($\approx$4.3 to $\approx$3.1) and a slow drift to 2.7--2.9 thereafter. The exception is instructive. Seed 3 tracks the other seeds through epoch 30 ($2.89$, mid-range) and destabilizes only in the final epochs, ending at $12.57$. The spike does mark real damage of one kind, since that seed's RankMe drops from the $\approx$185 band of the other seeds to $\approx$110, but its linear-probe accuracy is essentially unchanged (ur5 motor $R^2$ $0.334$, against $0.326$--$0.365$ for the healthy seeds). The three signals disagree because they measure different things. The loss reflects the joint state of the encoder and its predictor heads, RankMe the spectrum of the latent, and the probe the state recoverable from it; the loss is the least informative of the three about the exported encoder, which is why we evaluate with the encoder-only metrics defined next.

\subsubsection{Predictor-free performance metrics}
Training relies on predictor heads that map the fused latent to each held-out modality's target embedding, so the objective value measures the encoder and the predictors jointly. A collapsed latent paired with a predictor that has learned the collapse can score a near-perfect loss. We therefore evaluate the encoder with a family of measurements that involve no trained predictor. No single number in this family is treated as sufficient, because rank-style statistics alone can order models differently from downstream performance~\citep{jepamsac2026}; every table pairs them with probes.

\paragraph{Latent health (RankMe).} RankMe is the effective rank of the latent, the exponential of the entropy of the normalized singular-value spectrum~\citep{rankme2023}, bounded by $[1,\dim]$. It is a label-free collapse detector, and it catches a collapsed latent that a probe alone can miss. Because it scales with dimensionality, we report it alongside each feature's width and compare it only within a width family.

\paragraph{Latent organization (distance correlation and R\textsubscript{LDA}).} RankMe sees only the spectrum, so we add two measurements of how the latent is arranged. \emph{Distance correlation} samples pairs of held-out frames and reports the Spearman correlation between their latent distance and their true robot-state distance; a feature can be decodable by a probe yet carry no state structure in its geometry, and this statistic separates the two. \emph{R\textsubscript{LDA}}~\citep{jepamsac2026} is an LDA-style effective rank, the exponential entropy of the eigenvalues of the within-view versus between-sample covariance ratio, computed here with same-scene temporal neighbors as the views; it measures how many latent directions separate samples stably rather than how many carry variance, and it shares RankMe's dimensionality caveat.

\paragraph{Robot-state recovery (linear probe).} With the encoder frozen, we ask how much of the robot's true state can be read off a feature by a single linear map. The probe is ridge regression, that is, least-squares linear regression with an $\ell_2$ weight penalty (features standardized, $\alpha=10$), which stabilizes the fit when feature dimensions are correlated, as they are here. The probe is deliberately the weakest read-out available. If a linear map recovers a state variable, the information is not only present in the feature but arranged so that it is directly accessible, so the credit belongs to the representation rather than to the decoder; a high-capacity decoder would blur that attribution, and Section~\ref{sec:downstream} measures separately what a nonlinear read-out adds. We fit the probe on train rows and report $R^2\uparrow$ on held-out rows, where $1.0$ is perfect recovery, $0$ matches always predicting the mean, and negative values are worse than the mean. We score two target groups separately, \emph{motor} (joint angles and gripper) and \emph{end-effector} (force/torque and TCP pose). Unless noted, the probed feature is the \textbf{vision-only} fused latent $z_v$ with the state modalities masked at evaluation, so no robot state leaks into the probe input and the probe measures what the encoder learned to read out of pixels.

\subsubsection{Baselines}
\label{sec:baselines}
We compare the vision-only latent $z_v$ against three vision-only features, all probed under the identical protocol.
\begin{enumerate}
  \item \emph{Pretrained ViT (PT-ViT)}. Mean-pooled patch tokens (768-d) of the frozen LeJEPA ViT-B/16 pretrained on ImageNet-1k~\citep{lejepa2025}, the same backbone that feeds our encoder and the bar that fusion must beat.
  \item \emph{Pretrained ViT (PCA 256)}. The top 256 principal directions of the pretrained features, fit on train rows only. This is a pure compression control at the same width as $z_v$, so beating it rules out dimensionality reduction as the source of any gain.
  \item \emph{Pretrained ViT $+$ finetuned linear head}. The frozen backbone with a trained $768\!\rightarrow\!256$ projection, trained on our training split with the vision-only LeJEPA objective. This controls for lightweight in-domain adaptation at the same output width as $z_v$.
\end{enumerate}
The strictest control is a second encoder of identical architecture and size, trained on the same data with the state streams permanently masked. In the notation of \eqref{eq:encoder} its exported feature is $z_v^{\mathrm{vo}} = f_{\theta_{\mathrm{vo}}}(x_v, \emptyset, \emptyset)$, where the parameters $\theta_{\mathrm{vo}}$ never observe $x_m$ or $x_e$ during training; $z_v$ has the same functional form but its parameters $\theta$ were trained with all modalities present. This \emph{vision-only control} appears in the main table and is analyzed per embodiment in the ablation study (Section~\ref{sec:ablations}). One further baseline, a full finetune of the backbone, lowers every probe and changes no conclusion; it is described and reported in Appendix~\ref{sec:vitbase}.

\subsubsection{Quantitative results}
\label{sec:mainresults}
Table~\ref{tab:main} reports the 256-d features. The two 768-d baselines (pretrained ViT and finetuned ViT) are tabulated in Appendix~\ref{sec:vitbase} (Table~\ref{tab:vitbase}) under the identical protocol and are quoted below where relevant. Each cell averages over the three embodiments that carry both target groups (flexiv, ur5, kuka). Franka has no force/torque sensor and is degenerate for every feature (motor $R^2$ from $-0.35$ to $-6.71$), so it is reported in the per-embodiment break-out (Table~\ref{tab:phase1diag}) rather than folded into an average it would dominate.

\begin{table}[!htbp]
  \centering
  \small
  \caption{256-d features, averaged over the flexiv, ur5, and kuka held-out
  groups (franka is broken out in Table~\ref{tab:phase1diag}; the 768-d ViT baselines are in Table~\ref{tab:vitbase}). Linear probes read robot state on held-out rows from each frozen feature. For the trained encoders, $\pm$ is the std over 5 training seeds (training-seed uncertainty); for the remaining rows, which are deterministic given the split, $\pm$ is the cluster-bootstrap standard error over held-out groups (2000 replicates, test-set uncertainty). All four features are 256-d, so the RankMe values are directly comparable.}
  \label{tab:main}
  \begin{tabular}{lcr@{\,$\pm$\,}lcc}
    \toprule
    & & \multicolumn{2}{c}{} & \multicolumn{2}{c}{linear probe $R^2\uparrow$} \\
    \cmidrule(lr){5-6}
    feature & dim & \multicolumn{2}{c}{RankMe} & motor & end-effector \\
    \midrule
    \textbf{Kepler-Encoder-v0.1 ($z_v$)}       & 256 & 169.7 & 34.1 & 0.304 $\pm$0.019 & \textbf{0.282 $\pm$0.026} \\
    vision-only control ($z_v^{\mathrm{vo}}$)  & 256 & 189.9 & 2.5  & 0.198 $\pm$0.003 & 0.142 $\pm$0.003 \\
    pretrained ViT $+$ finetuned linear head   & 256 & 53.0  & 0.4  & 0.279 $\pm$0.008 & 0.206 $\pm$0.008 \\
    pretrained ViT (PCA 256)                   & 256 & 153.3 & 1.1  & \textbf{0.308 $\pm$0.010} & 0.234 $\pm$0.009 \\
    \bottomrule
  \end{tabular}
\end{table}

First, training with state raises what the encoder reads out of pixels. The fused $z_v$ and the vision-only control share architecture, data, and compute, and the fused latent probes higher on both target groups ($+0.106$ motor, $+0.140$ end-effector). The difference between the two runs is only whether state was present during training, so that is what the gap measures.

Second, the gain is specific to end-effector state, and within it, to force. On motor, the fused latent is statistically indistinguishable from both PCA 256 (paired cluster bootstrap over held-out groups, difference $-0.004$, 95\% CI $[-0.015, +0.009]$, $p=0.52$) and PT-ViT ($-0.008$, $p=0.37$); joint pose is largely visible to the camera, so vision-derived features already recover it. The motor tie is itself an average of per-embodiment differences in both directions, with the fused latent above PCA 256 on flexiv ($+0.033$, $p<0.001$) and below it on kuka ($-0.064$, $p<0.001$). On end-effector, the fused latent is above every baseline, and the gap is seed-robust and significant ($+0.049$ over PCA 256, 95\% CI $[+0.039, +0.060]$; $+0.062$ over PT-ViT; $p<0.001$ for every seed), holding on each sensored robot individually (Table~\ref{tab:phase1diag}). Isolating the force dimensions (Table~\ref{tab:forceonly}, Section~\ref{sec:ablations}), every vision-only feature reads force at or near zero (PT-ViT $-0.005$/$-0.115$/$0.103$ on flexiv/ur5/kuka), and the fused latent is above PT-ViT and the control on all three sensored robots (paired $t$, $p\le0.012$). Absolute force recovery is nonetheless modest ($0.05$/$-0.00$/$0.19$), consistent with a single timestep bounding what is recoverable (Appendix~\ref{sec:temporal}).

Third, in-domain adaptation of the backbone does not substitute for state, and rank does not predict probe accuracy. The finetuned linear head probes below PCA 256 on both targets while compressing to RankMe 53, the full finetune lowers both probes further (Table~\ref{tab:vitbase}), and force $R^2\approx0$ for every finetuned checkpoint (Appendix~\ref{sec:stage1}). A 5-seed retrain of the head reproduces this picture with negligible seed variance and slightly lower values (motor $0.258\pm0.002$, end-effector $0.190\pm0.001$, RankMe $44.2\pm0.0$), so the tabulated single run is the head's best case and its deficit is not a seed artifact. RankMe of $z_v$ sits near two thirds of the 256 ceiling, so the latent does not collapse; the vision-only control is higher-rank with weaker probes, so we report rank and probe accuracy together and read neither alone.

Fourth, the fused latent is organized by state, not merely decodable to it, and the organization comes from cross-modal training specifically. Table~\ref{tab:geometry} reports the two organization measurements of Section~\ref{sec:setup}. The distance correlation between latent distances and true state distances is $0.190$ (motor) and $0.221$ (end-effector) for the fused latent, against $|\rho|\le0.03$ for every frozen vision baseline and a near-flat $0.050$/$0.015$ for the vision-only control, whose single nonzero cell is kuka motor at $0.131$, half the fused latent's $0.241$ there. The fused latent is therefore the only feature whose geometry consistently tracks state, including on motor, where its probe accuracy is merely tied. Because the control shares the architecture, data, compute, and SIGReg regularization and differs only in never seeing state, the geometry is attributable to cross-modal training rather than to the architecture or the regularizer; the control is in fact the best-spread feature (RankMe $190$, Table~\ref{tab:main}) while carrying the least separable structure (R\textsubscript{LDA} $22.2$, below even PCA 256). R\textsubscript{LDA} gives the same ordering overall, with the fused latent highest ($75.6$) even against the 768-d PT-ViT ($34.9$), whose larger width favors it under this metric. Probes measure what a decoder can extract; these two measure how the space is arranged, and the baselines decode motor state without arranging their geometry by it.

\begin{table}[!htbp]
  \centering
  \small
  \caption{Latent organization, averaged over the flexiv, ur5, and kuka held-out
  rows (Kepler-Encoder-v0.1 and the vision-only control are 5-seed mean $\pm$std; the remaining baselines are deterministic given the split). Distance correlation ($d_{\mathrm{corr}}$) is the Spearman correlation between pairwise latent distance and pairwise true-state distance (20{,}000 pairs per embodiment; every Kepler correlation is individually significant at $p<10^{-17}$). R\textsubscript{LDA} follows \citet{jepamsac2026} with same-scene temporal neighbors as views; like RankMe it grows with width, which favors the 768-d PT-ViT.}
  \label{tab:geometry}
  \begin{tabular}{lcccc}
    \toprule
    feature & dim & $d_{\mathrm{corr}}$ motor & $d_{\mathrm{corr}}$ end-effector & R\textsubscript{LDA} \\
    \midrule
    \textbf{Kepler-Encoder-v0.1 ($z_v$)}     & 256 & \textbf{0.190 $\pm$0.067} & \textbf{0.221 $\pm$0.087} & \textbf{75.6 $\pm$5.6} \\
    vision-only control ($z_v^{\mathrm{vo}}$) & 256 & 0.050 $\pm$0.014 & 0.015 $\pm$0.015 & 22.2 $\pm$0.4 \\
    pretrained ViT $+$ finetuned linear head & 256 & 0.002 & $-$0.029 & 23.6 \\
    pretrained ViT (PCA 256)                 & 256 & 0.015 & $-$0.024 & 24.8 \\
    pretrained ViT (PT-ViT)                  & 768 & 0.020 & $-$0.020 & 34.9 \\
    \bottomrule
  \end{tabular}
\end{table}

The relative strength of PCA 256 has a simple reading. It is not a trained model but the variance-optimal linear compression of the PT-ViT features, so under a linear probe it inherits nearly all of their decodable content ($0.308$ vs.\ $0.315$ on motor). The two trained 256-d features are instead optimized for objectives other than preserving that content. The linear head is trained for multi-crop invariance under SIGReg; tracked across its training epochs, it sits at motor $0.275$--$0.280$ throughout, above a randomly initialized $768\!\rightarrow\!256$ projection ($0.251$) but never approaching PCA 256, while its spectrum collapses to RankMe 29 in the first epoch and recovers only to 53. Invariance training therefore recovers some decodable content over an arbitrary projection but preserves less of it than the variance-optimal one. Kepler-Encoder-v0.1 is trained for cross-modal prediction, which reshapes the code toward state-predictive structure and away from pixel-level appearance detail. On motor, where joint pose is visible in the frame and appearance detail is exactly what the probe uses, this leaves it tied with PCA 256 ($p=0.52$); on end-effector, where the pixels lack the information and preservation cannot supply it, the same trade puts the fused latent $0.049$ above ($p<0.001$). A feature that merely preserves the backbone is therefore the strongest 256-d baseline wherever the backbone already suffices, and fusion pays where it does not.

In sum, fusing state into training yields a moderate but consistent improvement in state recovery from pixels, and the improvement is concentrated where the camera carries the least information.

\subsubsection{Storage check: the latent retains its inputs}
\label{sec:storage}
The design excludes decoders from training and uses them as probes of what the latent stores (Section~\ref{sec:objective}); this section is the quantitative half of that check. We encode the \emph{full-context} latent $z$ (all modalities visible, as in training) and probe it for the same state targets on held-out rows. Table~\ref{tab:storage} shows the result. With state visible at encode time, a ridge probe alone reads back motor $R^2$ $0.57$ and end-effector $0.67$ (up to $0.84$ on kuka), roughly double the vision-only ceiling of Table~\ref{tab:main}, and the MLP raises end-effector to $0.73$. The bottleneck therefore retains a large share of what enters it after fusing three modalities into 256 dimensions, though not all of it. The generative decodes of Section~\ref{sec:qualitative} are the qualitative face of the same conclusion, recovering the scene from the vision path and arm pose from the state path.

\begin{table}[!htbp]
  \centering
  \small
  \caption{Storage check on the full-context latent $z$ (probes fit on train rows, $R^2\uparrow$ on held-out rows; 3-embodiment mean, 5-seed mean $\pm$std). The vision-only $z_v$ row repeats Table~\ref{tab:main} for reference. MLP inputs are clipped at $\pm10$ standard deviations because $z$ carries rare extreme activations on flexiv; the kuka motor MLP does not converge on some seeds, so the storage conclusion rests on the ridge row.}
  \label{tab:storage}
  \begin{tabular}{lcc}
    \toprule
    probe & motor $R^2\uparrow$ & end-effector $R^2\uparrow$ \\
    \midrule
    full-context $z$, ridge & 0.566 $\pm$0.039 & 0.674 $\pm$0.047 \\
    full-context $z$, MLP   & 0.431 $\pm$0.098 & 0.732 $\pm$0.028 \\
    vision-only $z_v$, ridge (reference) & 0.304 $\pm$0.019 & 0.282 $\pm$0.026 \\
    \bottomrule
  \end{tabular}
\end{table}

\subsubsection{Qualitative results}
\label{sec:qualitative}
\begin{figure}[!htbp]
  \centering
  \includegraphics[width=0.55\linewidth]{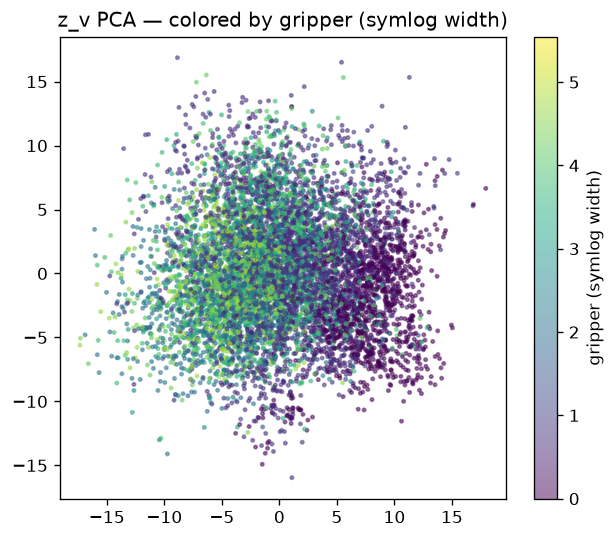}
  \caption{PCA of the vision-only $z_v$ within ur5, colored by gripper width, showing a smooth open$\rightarrow$closed gradient, i.e.\ the latent encodes continuous world-state, not only discrete robot/scene identity.}
  \label{fig:pcagripper}
\end{figure}

The latent organizes by continuous world-state. Within ur5, a PCA of the vision-only $z_v$ traces a smooth open-to-closed gripper gradient (Figure~\ref{fig:pcagripper}), so the latent encodes graded state rather than only robot or scene identity.

\begin{figure}[!htbp]
  \centering
  \begin{minipage}{\linewidth}
    \centering\includegraphics[width=\linewidth]{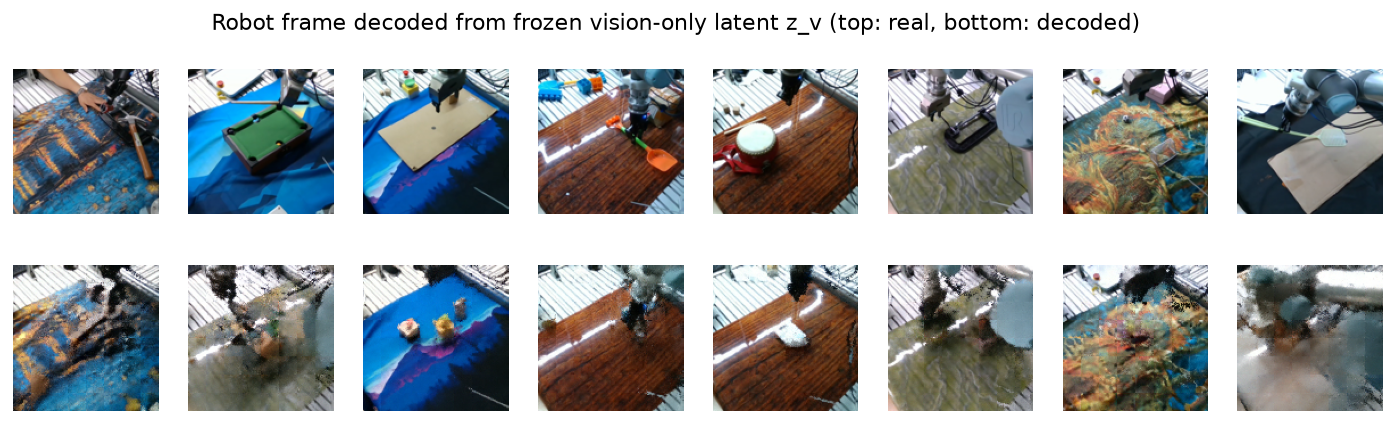}\\[2pt]
    {\small (a) Pixel decode from the vision-only latent $z_v$.}
  \end{minipage}\\[6pt]
  \begin{minipage}{\linewidth}
    \centering\includegraphics[width=\linewidth]{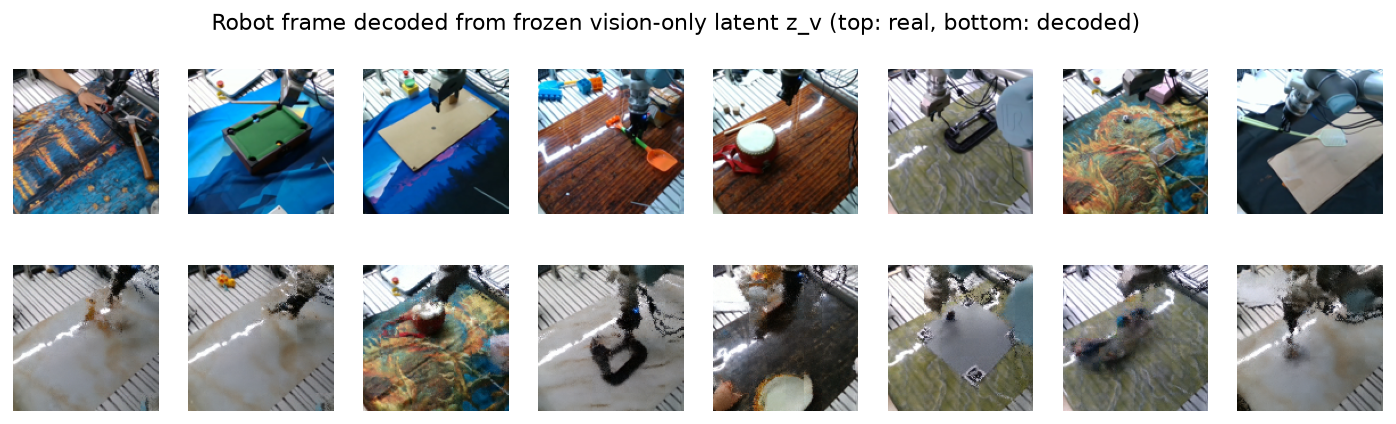}\\[2pt]
    {\small (b) Cross-modal decode from the state-only latent (vision hidden, motor$+$ee only).}
  \end{minipage}
  \caption{Generative probes of what the latent retains (held-out frames top, reconstructions bottom in each panel). \emph{(a)}~From the frozen vision-only $z_v$, PixNerd recovers recognizable scene layout, surfaces, and rough arm/object placement, but not fine texture, since the 256-d latent keeps world-state and drops pixel nuisance. \emph{(b)}~From the state-only latent, arm pose is recovered from proprioception while the surrounding scene is not, so the latent encodes the robot, not the world it cannot sense from state alone.}
  \label{fig:decode}
\end{figure}

Whereas a probe reports that information \emph{exists} in the latent, a generative decoder makes explicit what the latent represents, and decoding is one-to-many because the latent intentionally discards pixel nuisance (Section~\ref{sec:objective}). A small latent-conditioned diffusion decoder (PixNerd~\citep{pixnerd2026}) regenerates the camera frame from the frozen vision-only $z_v$ (Figure~\ref{fig:decode}a). Reconstructions are recognizable in scene layout, surfaces, colors, and rough arm and object placement, but blurry by design, because the 256-d latent keeps world-state and drops fine texture; quality tracks scene complexity, clean on simple scenes and degrading on clutter. The mirror experiment decodes from a \emph{state-only} latent (vision hidden, motor and ee only) and recovers arm pose from proprioception but not the surrounding scene (Figure~\ref{fig:decode}b). These reconstructions illustrate what the latent retains; the fusion claim rests on the quantitative comparison above and the ablations below.

Two further probes ask \emph{how} the vision path is reshaped to carry robot-relevant structure. The renderings are illustrative, and each is paired with the quantitative diagnostic that constrains its reading.

\begin{figure}[!htbp]
  \centering
  \includegraphics[width=\linewidth]{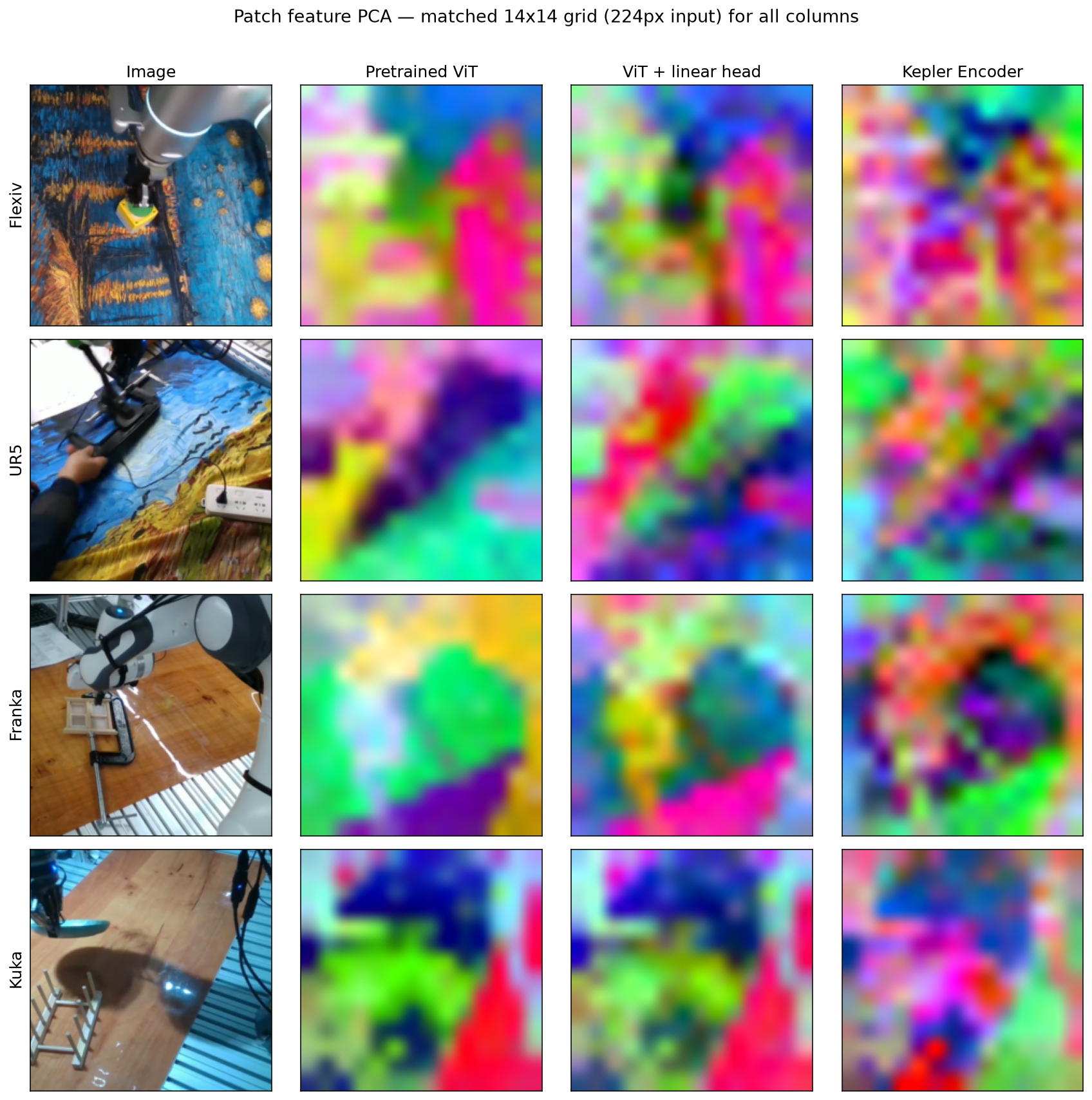}
  \caption{Patch-feature PCA over held-out test frames (rows: Flexiv, UR5, Franka,
  KUKA), all columns rendered at the matched $14\!\times\!14$ grid. \emph{Pretrained ViT} and \emph{ViT $+$ linear head} keep object-aligned regions; \emph{Kepler-Encoder-v0.1}'s vision projection (\emph{proj\_v}, the last spatial layer before cross-attention pooling) has the sharpest patch-to-patch transitions of the three but is less object-aligned, its colors fragmenting across object silhouettes. Each panel uses the same per-panel top-3 PCA-to-RGB reduction.}
  \label{fig:pcacompare}
\end{figure}

\paragraph{Patch-feature PCA.}
\label{sec:pca}
We compare three per-patch feature spaces over the same test-split frame at a matched $14\!\times\!14$ patch grid (224px input for every column), so no panel benefits from a finer rendering. Column~1 shows the frozen pretrained ViT (LeJEPA ViT-B/16~\citep{lejepa2025}) patch tokens. Column~2 shows the same frozen backbone with a single trained linear head (\emph{proj\_v}, $768\!\rightarrow\!256$; the pretrained ViT $+$ finetuned linear head of Section~\ref{sec:baselines}). Column~3 shows Kepler-Encoder-v0.1's own vision projection $\text{proj\_v}(\text{patch})+\text{mod}_v$, the last point in the pipeline where a spatial patch grid still exists, taken \emph{before} the fuser pools the patches into the non-spatial bottleneck, so it is the fair spatial comparison point. Each panel is reduced by the same recipe, its top three principal components with robust 2--98 percentile normalization, bicubic upsampled to the image, following the DINO/DINOv2 patch-PCA visualization convention~\citep{dinov2_2023}; the differing feature widths (768 vs.\ 256) rule out sharing one set of PCA axes across columns. Figure~\ref{fig:pcacompare} shows the result across four embodiments.

An earlier rendering of this comparison drew the frozen-ViT columns at a finer $28\!\times\!28$ grid and made the encoder projection appear smoother; at the matched grid that difference reverses, so the apparent smoothness was a resolution artifact. Quantitatively, the encoder projection has the sharpest patch-to-patch transitions of the three (mean adjacent-patch distance in the normalized PCA image $0.325$, against $0.246$ for the frozen ViT and $0.281$ for the linear head, averaged over the four scenes). What distinguishes the encoder is instead object alignment. The frozen ViT and the linear head render regions whose colors follow object silhouettes, with arm, workpiece, and background separating cleanly, while the encoder's principal-component colors fragment across those silhouettes, so its per-patch code is no longer organized primarily by local appearance. This reorganization is consistent with a code shaped for masked latent prediction rather than pixel fidelity (Section~\ref{sec:objective}, \citealp{ijepa2023}), though that link remains interpretive.

\begin{figure}[!htbp]
  \centering
  \includegraphics[width=\linewidth]{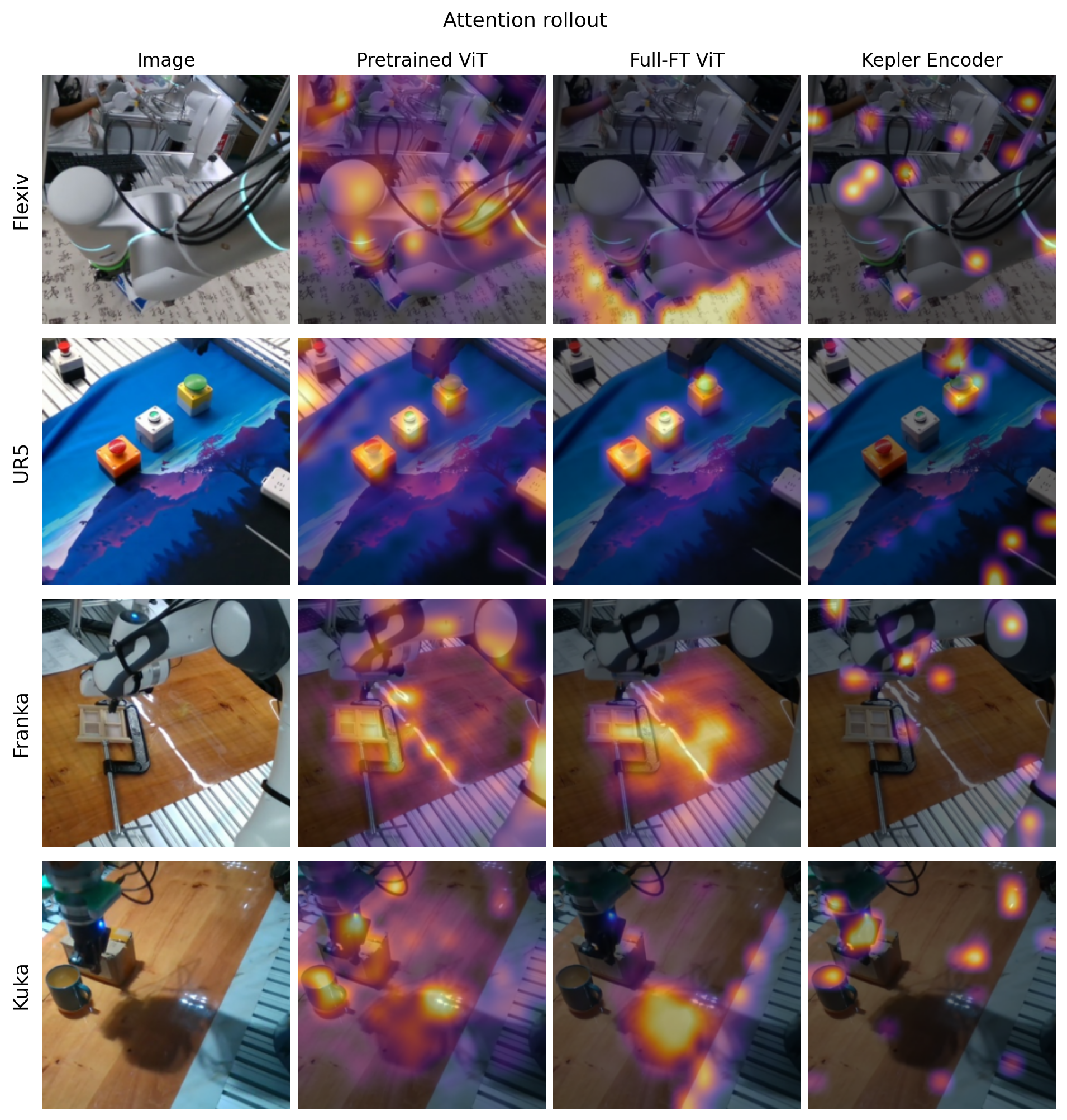}
  \caption{Attention overlays on held-out frames (rows: Flexiv, UR5, Franka,
  KUKA). \emph{Pretrained ViT} and \emph{Full-FT ViT} show CLS-to-patch self-attention; \emph{Kepler-Encoder-v0.1} shows the cross-attention bottleneck's attention over the 196 vision patches (heads and queries averaged). Attention sharpens from diffuse (pretrained) to compact regions on the manipulated object and gripper (encoder); a few off-action activations are hypothesized to anchor the scene frame (see text). Attention weight is illustrative, not a faithful importance measure~\citep{attnnotexpl2019}.}
  \label{fig:attncompare}
\end{figure}

\paragraph{Attention maps.}
\label{sec:attn}
The complementary question is spatial, asking where on the image each model's attention lands. We overlay three attention maps on the same frame sequence (Figure~\ref{fig:attncompare}). For the frozen pretrained ViT and a full-finetuned ViT (whole backbone trained on robot video), we take last-layer CLS-to-patch self-attention, averaged over heads (DINO-style~\citep{dinov2_2023}). For Kepler-Encoder-v0.1 we take the fuser's final cross-attention, the $\text{softmax}(qk^\top)$ weight of its bottleneck queries over the 196 vision patches (averaged over heads and queries), i.e.\ what the fused bottleneck \emph{reads} from the image. We note the standard caveat that raw attention weight is not a faithful importance attribution~\citep{attnnotexpl2019}; we use these maps as an illustrative complement to the probe results, not as proof.

The pretrained ViT's attention is diffuse, spread broadly over the scene with the generic, texture-driven saliency of backbones not adapted to the domain. Kepler-Encoder-v0.1's cross-attention is markedly more concentrated, collapsing onto a small number of compact regions, most of them on or near the manipulated object and the gripper, the regions where the manipulation occurs. Read alongside the cross-embodiment study (Section~\ref{sec:phase1}), this is the qualitative face of the same finding, in which the bottleneck has learned to select task-relevant image regions rather than average over the frame.

A few activations fall on regions where nothing is visibly happening. Two diagnostics constrain how to read them. First, they are largely not the known alternative in which ViTs park global information in a few high-norm background patches (register-token / attention-sink behavior~\citep{registers2023}). Over 96 held-out frames spanning 12 scenes, the cross-attention weights correlate only weakly with patch-token norms (mean Spearman $0.15$), and the top-10 attention patches overlap the top-10 high-norm patches at barely twice chance ($0.12$ against $0.05$). Second, the attention tracks scene content rather than fixed positions. Attention maps correlate at $0.36$ across frames within a scene but only $0.12$ across different scenes of the same config, whereas the high-norm patch positions themselves are strongly position-fixed across scenes ($0.45$). The off-action activations are therefore scene-dependent and not norm-driven, which is necessary for the \emph{anchoring} reading, in which the encoder attends to reference structures (table edges, fixtures, the robot base) that fix the geometric frame. Whether the attended regions are semantically meaningful landmarks remains unverified, and the moderate within-scene consistency shows the attention also moves as the manipulation progresses, so anchoring should be read as a tendency rather than a fixed set of points per scene.

\subsection{Ablation study}
\label{sec:ablations}
The main table averages over embodiments and pools force with pose. This section breaks both aggregations apart and isolates each ingredient of the result. All studies use 5 seeds unless noted.

\subsubsection{Cross-embodiment transfer}
\label{sec:phase1}
Using the robot-agnostic chunk packet (Section~\ref{sec:data}), we first ask whether the per-embodiment picture supports the averaged one, and whether one encoder trained on every robot matches per-robot specialists. Alongside Kepler-Encoder-v0.1 we train four specialists (flexiv $=$ cfg1$+$2, ur5 $=$ cfg3$+$4, franka $=$ cfg5, kuka $=$ cfg6$+$7), each for 5 seeds and 40 epochs, and probe every encoder on every embodiment's held-out groups. This yields a $5\times4$ transfer matrix; the probed feature is always the vision-only latent $z_v$ (state masked at eval), scored on the same two target groups as before. A single-embodiment proof-of-concept that preceded this study appears in Appendix~\ref{sec:stage2}. Table~\ref{tab:phase1diag} breaks the encoder against the two feature baselines out by embodiment, and Table~\ref{tab:phase1matrix} gives the specialists' full transfer matrix.

\begin{table}[!htbp]
  \centering
  \small
  \caption{Kepler-Encoder-v0.1 against the two feature baselines on each
  embodiment's held-out groups (vision-only $z_v$, 5-seed mean; seed standard deviations range $0.009$--$0.049$ and are omitted for readability). \emph{PT-ViT} (the pretrained ViT, 768-d mean-pooled frozen features) and \emph{PCA 256} (the top-256 principal directions of the PT-ViT features, a compression control) are fit on train rows only and deterministic given the split. The per-robot specialists appear in Table~\ref{tab:phase1matrix}. Franka (cfg5) has no F/T sensor, so it has no end-effector target.}
  \label{tab:phase1diag}
  \begin{tabular}{lcccccc}
    \toprule
    & \multicolumn{6}{c}{linear probe $R^2\uparrow$} \\
    \cmidrule(lr){2-7}
    & \multicolumn{3}{c}{motor (joints $+$ gripper)} & \multicolumn{3}{c}{end-effector (F/T $+$ pose)} \\
    \cmidrule(lr){2-4}\cmidrule(lr){5-7}
    robot & KEv0.1 $z_v$ & PT-ViT & PCA 256 & KEv0.1 $z_v$ & PT-ViT & PCA 256 \\
    \midrule
    flexiv & 0.252 & 0.232 & 0.220 & 0.268 & 0.211 & 0.205 \\
    ur5    & 0.339 & 0.321 & 0.321 & 0.187 & 0.103 & 0.144 \\
    kuka   & 0.321 & 0.391 & 0.384 & 0.393 & 0.353 & 0.353 \\
    franka & $-$0.377 & $-$6.709 & $-$0.750 & --- & --- & --- \\
    \bottomrule
  \end{tabular}
\end{table}

\begin{table}[!htbp]
  \centering
  \caption{Specialist transfer matrix, with each per-robot specialist (rows) probed on
  every robot's held-out groups (columns); vision-only $z_v$ $R^2\uparrow$, 5-seed mean. \textbf{Bold} $=$ diagonal (own robot). \emph{Left}: motor. \emph{Right}: end-effector (franka has no ee). Specialists degrade sharply off-diagonal; compare the per-embodiment KEv0.1 values in Table~\ref{tab:phase1diag}.}
  \label{tab:phase1matrix}
  \begin{minipage}{0.60\textwidth}
    \centering
    \small
    \begin{tabular}{lcccc}
      \toprule
      train $\downarrow$ / eval $\rightarrow$ & flexiv & ur5 & kuka & franka \\
      \midrule
      flexiv spec & \textbf{0.270} & 0.254 & 0.234 & $-$0.393 \\
      ur5 spec    & 0.148 & \textbf{0.324} & 0.236 & $-$0.411 \\
      kuka spec   & 0.121 & 0.221 & \textbf{0.330} & $-$0.431 \\
      franka spec & 0.069 & 0.088 & 0.148 & \textbf{$-$0.479} \\
      \bottomrule
    \end{tabular}
    \\[2pt]{\footnotesize motor (joints $+$ gripper)}
  \end{minipage}%
  \begin{minipage}{0.40\textwidth}
    \centering
    \small
    \begin{tabular}{lccc}
      \toprule
      & flexiv & ur5 & kuka \\
      \midrule
      flexiv spec & \textbf{0.286} & 0.137 & 0.247 \\
      ur5 spec    & 0.135 & \textbf{0.156} & 0.247 \\
      kuka spec   & 0.106 & 0.119 & \textbf{0.432} \\
      franka spec & 0.059 & 0.029 & 0.102 \\
      \bottomrule
    \end{tabular}
    \\[2pt]{\footnotesize end-effector (F/T $+$ pose)}
  \end{minipage}
\end{table}

\begin{table}[!htbp]
  \centering
  \caption{Latent health (RankMe of $z_v$, 5-seed mean $\pm$std), the exponential of the
  entropy of the normalized singular-value spectrum~\citep{rankme2023}, a label-free collapse detector measured on each robot's held-out $z_v$. RankMe has no learned baseline; it is bounded by $[1,\dim]$, with the ceiling ($256$, the latent width) meaning a perfectly uniform spectrum and the floor ($1$) full collapse. The KEv0.1 column's band ($\sim$165--180, about $0.65$--$0.70$ of ceiling) indicates no collapse; the specialist column sits slightly lower. Because RankMe scales with dimensionality, these values are comparable only within the 256-d $z_v$ family, never against the 768-d PT-ViT features. The KEv0.1 column is one encoder scored on four test sets, so its spread across robots reflects the test set, not four models.}
  \label{tab:phase1rankme}
  \begin{tabular}{lcc}
    \toprule
    robot & KEv0.1 $z_v$ & spec $z_v$ \\
    \midrule
    flexiv & 171.1 $\pm$32.6 & 161.5 $\pm$48.1 \\
    ur5    & 171.9 $\pm$31.4 & 177.1 $\pm$2.8 \\
    kuka   & 166.1 $\pm$27.8 & 142.4 $\pm$7.3 \\
    franka & 174.0 $\pm$30.9 & 146.6 $\pm$2.6 \\
    \bottomrule
  \end{tabular}
  \vspace{30pt}
\end{table}

\paragraph{The per-embodiment picture matches the average.} On the end-effector probe the Kepler-Encoder-v0.1 latent beats PT-ViT on every sensored robot ($+0.06$/$+0.08$/$+0.04$; Table~\ref{tab:phase1diag}). On motor the gain is partial. The latent beats PT-ViT on flexiv and ur5 but not on kuka, where joint pose is already visible to the camera, so we expect the motor payoff to come with the temporal extension (Appendix~\ref{sec:temporal}). Franka is the hardest embodiment, negative for every feature, but fusion stabilizes it ($-0.38$ against PT-ViT's $-6.71$).

\paragraph{One encoder covers four robots.} Kepler-Encoder-v0.1 matches per-robot specialists on their own robot (Table~\ref{tab:phase1diag} against the diagonal of Table~\ref{tab:phase1matrix}) and stays strong on every robot, while specialists degrade sharply off-diagonal (the franka specialist to $0.07$--$0.15$, even the strong kuka specialist to $0.11$--$0.22$; Table~\ref{tab:phase1matrix}). Two controls locate the gain. The pretrained ViT (PCA 256) stays within $0.01$--$0.04$ of PT-ViT, so the coverage is not a compression effect, and the data-budget-matched encoder below still beats every specialist off its training robot, so the coverage reflects embodiment diversity rather than data volume.

\paragraph{Per-embodiment latent health.} RankMe of $z_v$ holds a healthy $\sim$165--180 band on every test embodiment, about two thirds of the $256$ ceiling (Table~\ref{tab:phase1rankme}). Isolated low-rank seeds, including the loss-spike seed of Section~\ref{sec:setup}, pull some means down without moving the probes; these warrant a re-run rather than indicating a failure of the recipe.

\subsubsection{Cross-modal gain under matched compute}
\begin{figure}[!htbp]
  \centering
  \includegraphics[width=0.72\linewidth]{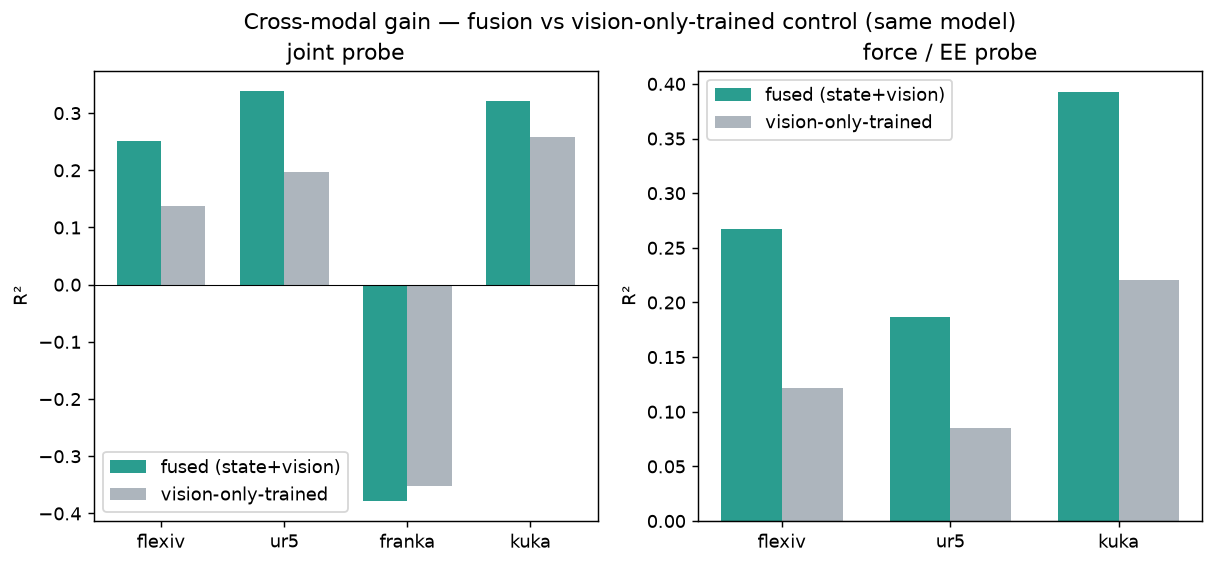}
  \caption{Cross-modal gain per robot, the fused $z_v$ against the identically-trained vision-only control. The gain is largest on the end-effector targets.}
  \label{fig:crossmodal}
\end{figure}

The vision-only control $z_v^{\mathrm{vo}}$ of Table~\ref{tab:main} is an identically-sized 256-d cross-attention encoder trained with the state streams permanently masked (same head, data, and compute; Section~\ref{sec:baselines}), which isolates cross-modal fusion from the alternative that we merely trained an in-domain encoder. Table~\ref{tab:crossmodal} breaks the comparison out on each robot's own held-out groups, and Figure~\ref{fig:crossmodal} plots the same comparison.

\begin{table}[h]
  \centering
  \small
  \caption{Cross-modal gain per embodiment ($R^2\uparrow$, 5-seed mean), comparing the fused latent $z_v$ with the compute-matched vision-only control $z_v^{\mathrm{vo}}$.}
  \label{tab:crossmodal}
  \begin{tabular}{lrrrr}
    \toprule
    & \multicolumn{2}{c}{motor $R^2$} & \multicolumn{2}{c}{end-effector $R^2$} \\
    \cmidrule(lr){2-3}\cmidrule(lr){4-5}
    robot & $z_v$ & $z_v^{\mathrm{vo}}$ & $z_v$ & $z_v^{\mathrm{vo}}$ \\
    \midrule
    flexiv & 0.252 & 0.137 & 0.268 & 0.122 \\
    ur5    & 0.339 & 0.198 & 0.187 & 0.085 \\
    kuka   & 0.321 & 0.258 & 0.393 & 0.220 \\
    franka & $-$0.377 & $-$0.352 & --- & --- \\
    \bottomrule
  \end{tabular}
\end{table}

\noindent Fusion adds $+0.06$--$0.14$ (motor) and $+0.10$--$0.17$ (end-effector) on the three data-rich robots. Because architecture and compute are identical, the gain is cross-modal learning rather than in-domain training; together with the pretrained ViT (PCA 256) control, this rules out dimensionality reduction as well. Franka is inconclusive, a small force-blind config where both variants are near-degenerate.

\subsubsection{Force isolated from TCP pose}
The end-effector probe mixes force/torque with TCP pose, so we split it (Kepler-Encoder-v0.1, 5-seed), probing force ($[0{:}6]$) and pose ($[6{:}15]$) separately. Pose-only $R^2$ is high ($0.41$/$0.31$/$0.53$ on flexiv/ur5/kuka) and dominates the aggregate, while force-only $R^2$ is modest ($0.05$/$-0.00$/$0.19$). The signal is nonetheless real and cross-modal. PT-ViT reads force at or near zero everywhere, whereas the fused latent is above it, and above the vision-only control $z_v^{\mathrm{vo}}$, on all three robots, with both gaps significant under a 5-seed paired $t$-test (Table~\ref{tab:forceonly}; $p\le0.012$). Partialling pose out of the force target leaves a near-zero residual for the fused latent ($+0.01$/$-0.03$/$+0.01$) and a negative one for PT-ViT, so the force recoverable at a single timestep is largely the component entangled with joint configuration and contact geometry; the pose-independent part is out of reach without time, motivating the temporal extension (Appendix~\ref{sec:temporal}).

\begin{table}[!htbp]
  \centering
  \small
  \caption{Force isolated from pose ($R^2\uparrow$ on held-out rows; the $z_v$ force column is 5-seed mean$\pm$std, the remaining columns are 5-seed means or deterministic). Force $=$ F/T dims $[0{:}6]$. The fused latent beats PT-ViT and the compute-matched vision-only control $z_v^{\mathrm{vo}}$ on every sensored robot (paired $t$, $p\le0.012$), though absolute force $R^2$ is modest; pose-only and the pose-partialled residual (force$\perp$pose) are shown for context.}
  \label{tab:forceonly}
  \begin{tabular}{lr@{\,$\pm$\,}lrrrr}
    \toprule
    robot & \multicolumn{2}{c}{$z_v$ force} & $z_v^{\mathrm{vo}}$ force & PT-ViT force & $z_v$ pose & $z_v$ force$\perp$pose \\
    \midrule
    flexiv & $0.049$ & $0.009$ & $0.010$ & $-0.005$ & $0.413$ & $+0.008$ \\
    ur5    & $-0.001$ & $0.008$ & $-0.019$ & $-0.115$ & $0.311$ & $-0.025$ \\
    kuka   & $0.187$ & $0.021$ & $0.067$ & $0.103$ & $0.530$ & $+0.007$ \\
    \bottomrule
  \end{tabular}
\end{table}

\subsubsection{Data-budget-matched Kepler-Encoder-v0.1}
To separate embodiment diversity from data volume, we retrain Kepler-Encoder-v0.1 on a random $32{,}870$-frame subset (the mean specialist's training size; same $40$ epochs, hence a matched gradient-step budget), 5 seeds. Matched to a single specialist's budget, the encoder falls $0.02$--$0.06$ below each specialist on that specialist's own robot (Table~\ref{tab:matched}), so the full model's on-diagonal parity partly reflects its $4\times$ larger training set. The embodiment-agnostic result, however, is not a volume artifact. At the same total budget the shared encoder covers all four robots (motor $R^2$ $0.22$/$0.30$/$0.31$ on flexiv/ur5/kuka) and beats every specialist evaluated \emph{off} its training robot ($0.07$--$0.25$; Table~\ref{tab:phase1matrix}). Distributing a fixed data budget across embodiments buys broad competence that concentrating it on one does not.

\begin{table}[!htbp]
  \centering
  \small
  \caption{Data-budget-matched Kepler-Encoder-v0.1 (vision-only $z_v$ motor $R^2\uparrow$, 5-seed mean). ``matched KEv0.1'' is trained on $32{,}870$ frames, the mean specialist's budget. It trails the specialist on-diagonal by $0.02$--$0.06$ but, unlike any specialist, stays strong across all robots.}
  \label{tab:matched}
  \begin{tabular}{lrrrr}
    \toprule
    robot & matched KEv0.1 & full KEv0.1 & specialist & PT-ViT \\
    \midrule
    flexiv & $0.215$ & $0.252$ & $0.270$ & $0.232$ \\
    ur5    & $0.295$ & $0.339$ & $0.324$ & $0.321$ \\
    kuka   & $0.308$ & $0.321$ & $0.330$ & $0.391$ \\
    franka & $-0.390$ & $-0.377$ & $-0.479$ & $-6.709$ \\
    \bottomrule
  \end{tabular}
\end{table}

\subsection{Downstream capabilities of the frozen encoder}
\label{sec:downstream}
Beyond probes, we ask what the \emph{frozen} encoder is directly useful for, with no retraining. These results show the latent supports concrete downstream use; the fusion claim itself rests on the quantitative comparison and the ablation study above, not on the demonstrations here.

\begin{figure}[!htbp]
  \centering
  \includegraphics[width=0.8\linewidth]{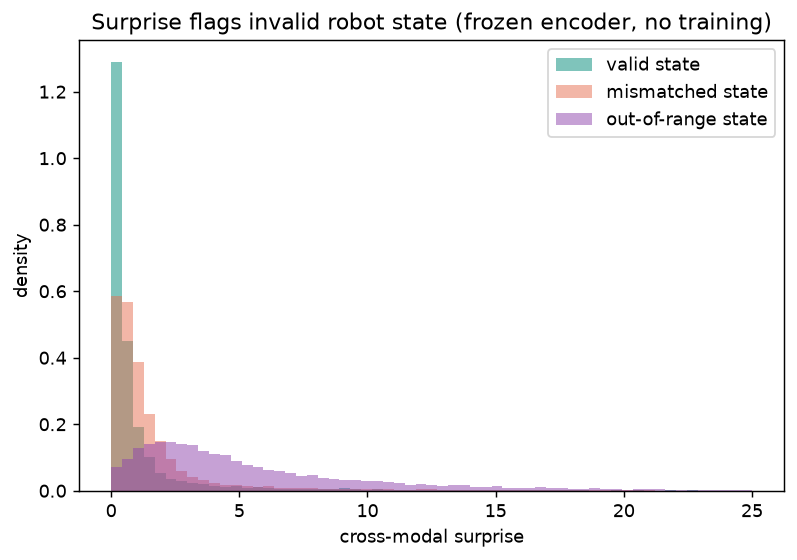}
  \caption{Invalid-state detection on held-out ur5. The encoder's cross-modal prediction error (``surprise'') separates valid robot state from corrupted state with no additional training.}
  \label{fig:surprise}
\end{figure}

\paragraph{Invalid-state and safety detection.} Because the encoder is trained by cross-modal prediction, its per-sample prediction error is a built-in \emph{surprise} signal, low when the robot state is consistent with vision and high when it is not. On ur5 held-out, valid vs.\ corrupted state separates at AUROC $0.90$ (out-of-range state) and $0.69$ (state swapped from another scene), with no additional training (Figure~\ref{fig:surprise}), giving a robot-safety anomaly detector on the encoder we already have.

\begin{figure}[!b]
  \centering
  \includegraphics[width=\linewidth]{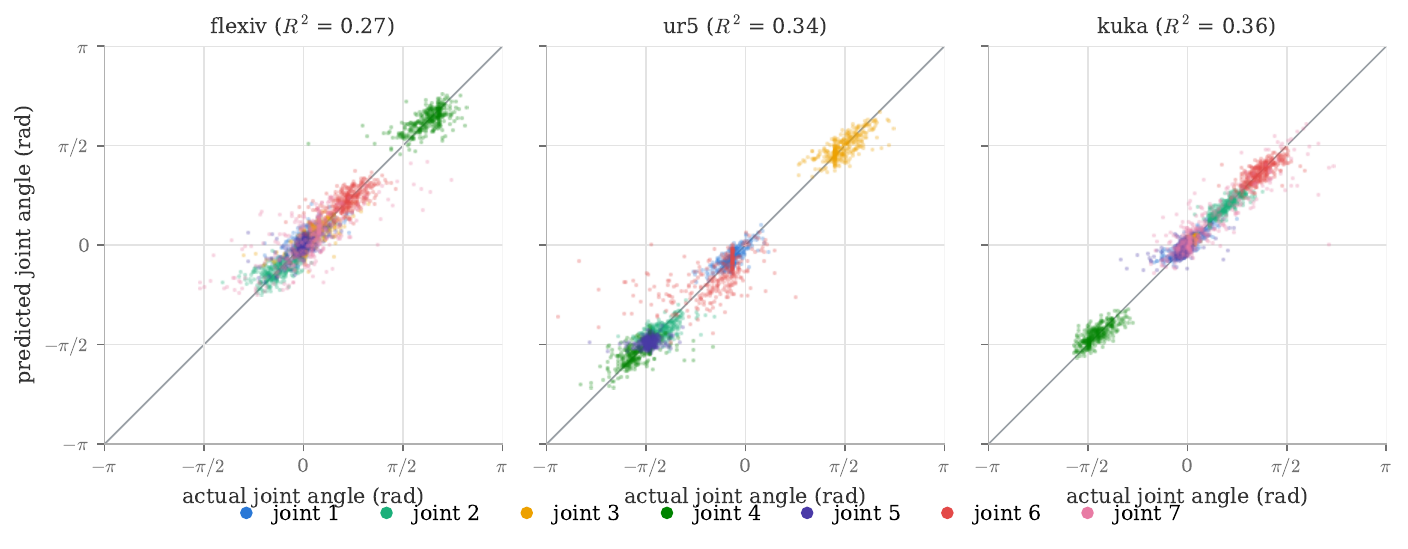}
  \caption{Joint angles decoded from the frozen vision-only $z_v$ by the standard ridge probe (seed 0), predicted vs.\ actual on each robot's held-out rows, all joints overlaid (300 points per joint; angles recovered via $\mathrm{atan2}$ from the predicted $\sin$/$\cos$ channels). Panel titles give each robot's overall motor $R^2$. The legend is shared across panels; its joint-7 entry applies only to the 7-DOF flexiv and kuka panels, while the ur5 panel shows 6 joints, since the grid's seventh joint row is unpopulated for a 6-DOF arm.}
  \label{fig:statedecode}
\end{figure}

\paragraph{State read-out.} The linear probe of Section~\ref{sec:setup} lower-bounds what a feature contains, because it can only read information that is linearly arranged. To measure the headroom above that bound we fit MLP read-outs to held-out ur5 motor state, identically for every feature, and the decoder's own training budget turns out to matter. An under-trained MLP (two hidden layers of 512, early stopping, at most 200 iterations) shows no gain on $z_v$ ($0.328\pm0.020$ against ridge $0.339\pm0.014$); the same architecture trained to convergence (300 epochs, no early stopping) lifts $z_v$ to $0.441\pm0.020$. Three conclusions follow at full decoder strength. First, the 256-d latent stores motor state beyond the linear bound ($+0.10$ over ridge). Second, the fusion gain persists, since the vision-only control reaches only $0.315\pm0.005$ under the same decoder. Third, preservation still wins on motor, with PT-ViT at $0.523$ and PCA 256 at $0.459$, so compression keeps its motor-state price at high decoder capacity, consistent with the motor tie of Section~\ref{sec:mainresults}. Figure~\ref{fig:statedecode} shows the decode itself, with joint angles recovered from the ridge probe's predicted $\sin$/$\cos$ channels on each robot's held-out rows, the paper's standard read-out. Gripper aperture behaves differently. Its distribution is near-discrete, mostly fully open or fully closed, so we score it as classification; $z_v$ gives $78\%$ accuracy and $0.87$ AUROC against $77\%$ and $0.86$ for PT-ViT, a near-tie consistent with gripper state being visible in pixels.

\paragraph{Read-outs are configuration-specific.} A natural deployment hope is to fit one read-out on the fully-sensored latent and reuse it when sensors drop out. Table~\ref{tab:transfer} shows this does not work here. The same ridge probe that reads the full-context latent $z$ at $0.57$/$0.67$ scores \emph{below zero} when applied unchanged to $z_v$, worse than predicting the mean on every embodiment and target, while refitting on $z_v$ recovers the Table~\ref{tab:main} performance; the MLP read-out transfers no better. The two latents therefore encode state along different directions. This is consistent with the training objective, which predicts the held-out modality's embedding through a per-modality head (Section~\ref{sec:objective}) and never aligns $z_v$ with $z$ itself. The practical rule for a downstream consumer at a single timestep is to fit its read-out on the latent of the sensor configuration it will run with; what stays invariant across configurations is the encoder, not the probe. The finding also motivates the temporal extension (Appendix~\ref{sec:temporal}). In deployment, sensor dropout is usually transient, and a sensor that vanishes mid-stream remains constrained by its own recent past. An encoder that fuses across time can therefore hold the latent in one coordinate frame through the dropout, and whether that restores read-out transfer is a directly testable prediction for the temporal model.

\begin{table}[!b]
  \centering
  \small
  \caption{Probe transfer across sensor configurations (ridge probes, $R^2\uparrow$ on held-out rows; 3-embodiment mean, 5-seed mean $\pm$std). A probe fit on the full-context latent $z$ collapses when applied unchanged to the vision-only latent $z_v$, while refitting on $z_v$ recovers the main-table performance.}
  \label{tab:transfer}
  \begin{tabular}{lr@{\,$\pm$\,}lr@{\,$\pm$\,}l}
    \toprule
    & \multicolumn{4}{c}{linear probe $R^2\uparrow$} \\
    \cmidrule(lr){2-5}
    read-out & \multicolumn{2}{c}{motor} & \multicolumn{2}{c}{end-effector} \\
    \midrule
    fit on $z$, evaluate on $z$ (storage, Section~\ref{sec:storage}) & 0.566 & 0.039 & 0.674 & 0.047 \\
    fit on $z$, evaluate on $z_v$ (transferred)               & $-$0.628 & 0.431 & $-$0.534 & 0.269 \\
    fit on $z_v$, evaluate on $z_v$ (refit, Table~\ref{tab:main}) & 0.304 & 0.019 & 0.282 & 0.026 \\
    \bottomrule
  \end{tabular}
\end{table}

\section{Discussion and Limitations}
\label{sec:discussion}
\paragraph{An information-flow reading of the results.} Three numbers summarize what the encoder does with state. State that enters is largely kept, since a ridge probe reads motor $R^2$ $0.57$ and end-effector $0.67$ back out of the full-context latent (Section~\ref{sec:storage}). State that is absent at evaluation is partially inferred from pixels, at $0.30$ and $0.28$ from the vision-only latent, with the gain over vision baselines significant only where the camera is weakest, on end-effector state and force in particular (Section~\ref{sec:mainresults}). And the two access paths are geometrically distinct, since a probe fit on one latent fails on the other (Section~\ref{sec:downstream}). Read alongside the organization results (Table~\ref{tab:geometry}), where the fused latent is the only feature whose distances track state, the picture is of an encoder that reorganizes the vision path around body state rather than one that appends state to appearance features.

\paragraph{Single-timestep.} Every result here fuses modalities at a single tick, so the encoder captures structure \emph{across modalities} but not \emph{across time}. True time-derivatives (velocity, acceleration) and contact dynamics are out of reach, which also bounds the modest force recovery; the native-rate temporal extension (Appendix~\ref{sec:temporal}) is where we expect the motor and force payoff.

\paragraph{Correlational, not causal.} Same-time alignment shows the latent \emph{contains} cross-modal structure, not that it models dynamics; action-conditioned forward prediction, the causal step, is future work (Appendix~\ref{sec:temporal}). The evaluation is probe-centric and aligned with the training target, which is favorable to the method, but three safeguards bound the risk. The controls of Section~\ref{sec:ablations} rule out in-domain training and compression, the paired cluster bootstrap of Section~\ref{sec:mainresults} marks the motor comparison as a tie rather than a win, and the predictor-free measurements of Section~\ref{sec:setup} (RankMe, distance correlation, R\textsubscript{LDA}) involve no fitted decoder at all.

\paragraph{No zero-shot transfer to unseen robots.} Embodiment-agnosticism holds for robots \emph{in the training mix} (Section~\ref{sec:phase1}), not for unseen ones. In a leave-one-robot-out test the zero-shot latent trails PT-ViT on the held-out robot's motor probe by $0.03$--$0.13$ (Table~\ref{tab:loro}), as the fused vision path tunes to the training embodiments' visual domains. It is, however, far more stable than PT-ViT on the degenerate franka config and beats it on the ur5 end-effector probe. Few-shot adaptation and a broader embodiment set are future work.

\begin{table}[!htbp]
  \centering
  \small
  \caption{Zero-shot leave-one-robot-out (vision-only $z_v$ $R^2\uparrow$, 5-seed mean): each robot is probed by an encoder trained on the \emph{other three} and never exposed to it. The zero-shot latent trails PT-ViT on motor for every non-degenerate robot; franka is force-blind and degenerate for both (PT-ViT collapses to $-6.71$).}
  \label{tab:loro}
  \begin{tabular}{lrrrr}
    \toprule
    held-out robot & LORO $z_v$ motor & PT-ViT motor & LORO $z_v$ ee & PT-ViT ee \\
    \midrule
    flexiv & $0.143$ & $0.232$ & $0.131$ & $0.211$ \\
    ur5    & $0.295$ & $0.321$ & $0.161$ & $0.103$ \\
    kuka   & $0.259$ & $0.391$ & $0.269$ & $0.353$ \\
    franka & $-0.459$ & $-6.709$ & --- & --- \\
    \bottomrule
  \end{tabular}
\end{table}

\paragraph{Read-outs do not transfer across sensor configurations.} A probe fit on the full-context latent does not apply to the vision-only latent (Table~\ref{tab:transfer}), so a consumer must fit its read-out for the sensor configuration it deploys with. The objective never asks the two latents to agree, only to predict each other's modality embeddings (Section~\ref{sec:objective}), so this is a designed property with a real cost. Two extensions could recover single-read-out deployment. A latent-consistency term aligning $z_v$ with $z$ is the direct route, and its interaction with the predict-do-not-equate principle is an open design question. The temporal extension (Appendix~\ref{sec:temporal}) is the indirect route, since real sensor dropout is transient and a temporally fused latent remains constrained by the dropped sensor's recent past; whether temporal fusion restores read-out transfer is a testable prediction of that model. Relatedly, compression has a price at high decoder capacity, since an MLP extracts more motor state from the raw 768-d features than from the 256-d latent (Section~\ref{sec:downstream}).

\paragraph{Other caveats.} A single external camera is used (multi-view is future work); RH20T's shipped streams are camera-aligned rather than native-rate, so the multi-rate design is exercised against the raw high-frequency logs; and cfg5's force is a robot-internal estimate rather than a physical sensor reading.

\section{Conclusion}
\label{sec:conclusion}
Kepler-Encoder-v0.1 fuses video, proprioception, and force/torque into a single latent with a learned-query cross-attention layer, trained by masked cross-modal latent prediction under LeJEPA/SIGReg, non-collapsing, prediction-based, and robust to dropped or hardware-absent sensors by construction. Our question was whether this buys anything over vision alone. On RH20T the answer is yes where the camera is weakest. The vision-only latent carries end-effector state, and force in particular, that raw frozen-ViT features do not, and it is the only feature whose latent geometry tracks state; on motor state, which the camera largely sees, it is statistically tied with the strongest vision baselines, and two controls rule out in-domain training and mere compression. One embodiment-agnostic encoder covers the four training robots, competitive with per-robot specialists and, at a matched data budget, still covering robots specialists cannot, so its breadth reflects embodiment diversity, not data volume; zero-shot transfer to unseen robots remains open. The frozen latent is directly useful. Its own cross-modal prediction error is a training-free monitor for invalid robot states, a natural hook for latent-level safety intervention, and a diffusion decoder reconstructs the camera frame from it, confirming the spatial compression preserves world-state. The clearest limitation is that everything here is single-timestep. The natural next steps are to fuse streams across time at their native rates (Appendix~\ref{sec:temporal}), which would also separate this encoder from robot self-supervised methods that assume synchronized observations and may restore read-out transfer under sensor dropout, and to build an action predictor on the frozen latent, the direction we expect to be most useful.

\clearpage
\bibliographystyle{plainnat}
\bibliography{references}

\appendix
\clearpage

\section{Preliminary: freezing the vision backbone}
\label{sec:stage1}
Before the main study, one preliminary experiment fixes a design choice, whether to finetune the vision backbone on robot video. It does not help. Continuing LeJEPA on cfg3 video has no effect. A high learning rate (2e-4) collapses the encoder (RankMe $300\!\rightarrow\!158$); a gentle rate (2e-5) restores health (RankMe $\sim$285) but leaves robot-relevant probes flat (contact $0.69\!\rightarrow\!0.67$), and force $R^2\approx0$ for every checkpoint. The apparent drop in a task-id probe ($0.91\!\rightarrow\!0.74$) is a \emph{saturated} metric tracking ImageNet appearance, not a real regression, a first instance of over-reliance on a single metric. The lasting finding motivates the whole model. \textbf{A single frame does not contain force}, so vision alone cannot encode it. Hence the backbone is frozen (warm-started from the pretrained checkpoint) and force must enter through a different modality.

\clearpage

\section{Frozen and finetuned ViT baselines}
\label{sec:vitbase}
Table~\ref{tab:vitbase} reports the two 768-d ViT baselines, evaluated under the protocol of Section~\ref{sec:mainresults} (same splits, targets, and ridge probes; averages over the flexiv, ur5, and kuka held-out groups). \emph{PT-ViT} is the frozen pretrained backbone of Section~\ref{sec:baselines}. The \emph{finetuned ViT} is a full finetune of that backbone on our training split with the vision-only LeJEPA objective, at the highest learning rate that does not collapse the encoder ($2\times10^{-5}$; Appendix~\ref{sec:stage1}), evaluated as a single run. The pretrained features set the strongest motor baseline ($0.315$). The full finetune lowers both probes relative to the frozen backbone; this result, together with the head-only finetune and the PCA 256 compression of Table~\ref{tab:main}, shows that in-domain adaptation of the backbone does not improve state recovery, in line with the freezing decision of Appendix~\ref{sec:stage1}.

\begin{table}[!htbp]
  \centering
  \small
  \caption{The 768-d ViT baselines, averaged over the flexiv, ur5, and kuka
  held-out groups. The pretrained ViT is deterministic given the split; its $\pm$ is the cluster-bootstrap standard error over held-out groups. The finetuned ViT is a single run without a computed interval. RankMe scales with dimensionality and is comparable only within a width family, so these values are not comparable to the 256-d family of Table~\ref{tab:main}.}
  \label{tab:vitbase}
  \begin{tabular}{lcccc}
    \toprule
    & & & \multicolumn{2}{c}{linear probe $R^2\uparrow$} \\
    \cmidrule(lr){4-5}
    feature & dim & RankMe & motor & end-effector \\
    \midrule
    pretrained ViT (PT-ViT)   & 768 & 380.7 $\pm$2.8 & 0.315 $\pm$0.013 & 0.222 $\pm$0.014 \\
    finetuned ViT             & 768 & 252.0 & 0.233 & 0.158 \\
    \bottomrule
  \end{tabular}
\end{table}

\clearpage

\section{Notation}
\label{sec:notation}
Table~\ref{tab:notation} lists every symbol used in the Method (Section~\ref{sec:method}), together with the evaluation-specific symbols introduced in Section~\ref{sec:experiments}. Symbols are chosen so that no glyph carries two meanings. Two conventions resolve near-collisions. Case is significant, so $M$ (query count) and $m$ (the motor modality) are distinct, as are $N$ (a token count) and any lowercase index. Weight is significant, so a plain $e$ (the ee modality label) and a boldface $\mathbf{e}_{(\cdot)}$ (a learned modality embedding) are distinct. The generic held-out modality is always indexed by $k$, never by $m$, so $m$ refers only to the motor stream.

\begin{table}[!htbp]
  \centering
  \small
  \begin{tabular}{@{}ll@{}}
    \toprule
    \textbf{Symbol} & \textbf{Meaning} \\
    \midrule
    \multicolumn{2}{@{}l}{\emph{Modalities and streams}} \\
    $v,\,m,\,e$ & modality labels, vision / motor / end-effector (ee) \\
    $k$ & generic modality index, $k\in\{v,m,e\}$ (used for held-out terms) \\
    $x_v,\,x_m,\,x_e$ & raw input of each stream (RGB frame, motor grid, ee window) \\
    $F_v$ & frozen ViT patch features of the frame $x_v$ ($N_v \times 768$) \\
    $\emptyset$ & a dropped stream at evaluation, no input provided \eqref{eq:encoder} \\
    $\mathbf{v}_m$ & motor validity mask, matching $x_m$ in shape (1 $=$ present) \\
    $\phi_m$ & motor featurizer, zeroes invalid channels and appends the mask bits \\
    $q,\,\dot q$ & joint angle and joint velocity (motor channel encodings) \\
    \midrule
    \multicolumn{2}{@{}l}{\emph{Tokens and dimensions}} \\
    $d$ & shared token / latent width \\
    $N_v,\,N_m,\,N_e$ & per-stream token counts; $N = N_v + N_m + N_e$ total \\
    $W_v,\,W_m,\,W_e$ & per-stream linear projections to width $d$ \\
    $\mathbf{e}_{(\cdot)}$ & learned modality embedding (bold, distinct from label $e$) \\
    $\mathbf{p}_m,\,\mathbf{p}_e$ & learned position embeddings for the structured streams \\
    $T_v,\,T_m,\,T_e$ & projected, tagged tokens of each stream \\
    $C$ & context set, $C\in\mathbb{R}^{N\times d}$, the concatenation of all tokens \\
    \midrule
    \multicolumn{2}{@{}l}{\emph{Fuser}} \\
    $M$ & number of learned queries (capital, distinct from motor $m$) \\
    $\xi_i$ & the $i$-th learned query vector; $Q^{(\ell)}$ the query set after block $\ell$ \\
    $L$ & fuser depth (number of cross-attention blocks) \\
    $\ell$ & block index, $\ell = 1,\dots,L$ \\
    $\mathcal{M}$, $\mathcal{M}_{\mathrm{valid}}$ & attention block-mask; the valid variant blocks only invalid/padding tokens \\
    $\mathcal{M}_{\mathrm{hide}(S)}$ & mask additionally blocking every modality in $S$ \\
    $\mathrm{Fuse}$, $\mathrm{CrossAttn}$, $\mathrm{FFN}$, $\mathrm{LN}$ & fuser, cross-attention, feed-forward, and layer-norm maps \\
    \midrule
    \multicolumn{2}{@{}l}{\emph{Latents and objective}} \\
    $z$ & fused latent from the full context \\
    $z^{(\backslash k)}$ & fused latent with modality $k$ held out \\
    $z_v$ & vision-only latent (motor and ee hidden), the exported embedding \\
    $z_v^{\mathrm{vo}}$ & latent of the vision-only-\emph{trained} control encoder $f_{\theta_{\mathrm{vo}}}$ (Section~\ref{sec:baselines}) \\
    $d_{\mathrm{corr}}$ & Spearman correlation between pairwise latent and true-state distances (Section~\ref{sec:setup}) \\
    R\textsubscript{LDA} & LDA effective rank over temporal-neighbor views~\citep{jepamsac2026} (Section~\ref{sec:setup}) \\
    $f_\theta,\,\bar f_{\bar\theta}$ & online encoder and its EMA target copy \\
    $h_k$ & per-modality predictor head for modality $k$ \\
    $\operatorname{sg}[\cdot]$ & stop-gradient operator \\
    $\mathcal{L}_{\mathrm{pred}}$ & masked cross-modal prediction loss \\
    \bottomrule
  \end{tabular}
  \caption{Symbols used in the Method (Section~\ref{sec:method}) and the evaluation (Section~\ref{sec:experiments}).}
  \label{tab:notation}
\end{table}

\clearpage

\section{Single-embodiment fusion: cfg3 POC and cfg3\texorpdfstring{$+$}{+}cfg4 scale-up}
\label{sec:stage2}
Before scaling to the full transfer matrix (Section~\ref{sec:phase1}), we fuse video $+$ robot state at a single timestep on UR5 data and ask whether the fused latent beats each modality alone, and whether the gain is \emph{cross-modal} rather than mere compression. A cross-modal signal gate confirms the premise (frozen vision $\rightarrow$ state $R^2=0.43$, TCP $0.73$), justifying the fusion encoder. We train the cross-attention fuser with masked cross-modal latent prediction $+$ per-modal $+$ joint SIGReg, and at evaluation feed \emph{vision only}, probing to robot state. Table~\ref{tab:stage2} reports two runs, a cfg3 proof-of-concept ($\sim$24k frames) and a cfg3$+$cfg4 scale-up (86{,}430 frames / 2{,}881 scenes, $3.6\times$ the data).

\begin{table}[!htbp]
  \centering
  \caption{Single-embodiment fusion (UR5). Predicting held-out robot state from
  each feature ($R^2\uparrow$, 5 seeds, group-held-out, encoder retrained per split). The cross-modal latent $z_v$ (vision-only at eval) beats both raw vision and the PCA compression control on all 5 seeds in both runs. RankMe of $z_v$ is $\sim$211 (ceiling 256), higher than the $\sim$165--180 cross-embodiment band of Section~\ref{sec:phase1} because this latent is scored on a single embodiment rather than averaged over four heterogeneous test sets; both indicate no collapse.}
  \label{tab:stage2}
  \begin{tabular}{lcc}
    \toprule
    feature $\rightarrow$ predict robot state & cfg3 POC & cfg3$+$cfg4 \\
    \midrule
    \textbf{Fused $z_v$ (256-d, cross-modal, state masked)} & \textbf{0.551 $\pm$0.018} & \textbf{0.653 $\pm$0.008} \\
    PT-ViT, raw vision (768-d, mean-pooled)                     & 0.257 $\pm$0.075          & 0.516 $\pm$0.010 \\
    pretrained ViT (PCA 256, compression control)              & 0.134 $\pm$0.047          & 0.418 $\pm$0.015 \\
    \bottomrule
  \end{tabular}
\end{table}

$z_v$ beats raw vision by $+0.294\pm0.078$ (POC) / $+0.137\pm0.007$ (scale) and the PCA control by $+0.417\pm0.050$ / $+0.235\pm0.010$, positive on every seed. Two facts make this meaningful. First, \textbf{beating the PCA control shows the gain is cross-modal, not compression}, because a latent $3\times$ smaller than raw vision predicts the robot \emph{better}, while PCA 256, the same compression with no cross-modal training, loses. Second, \textbf{only vision enters at eval}, so the encoder has \emph{learned to read robot-relevant structure out of pixels} by having been trained alongside state. This is exactly what a late-fusion recipe (train each modality's encoder separately, then concatenate) cannot produce, because its vision features were never shaped by force or joints.

\paragraph{Why every baseline rises from cfg3 to cfg3$+$cfg4.}
The margin over raw narrows in absolute terms ($+0.294\!\rightarrow\!+0.137$) but stays decisive over the compression control ($+0.417\!\rightarrow\!+0.235$). The baselines rise largely for reasons that are \emph{not} ``raw vision got better at decoding within-robot state.'' (i)~$R^2$'s denominator changed, since pooling two UR5 setups injects easy, pixel-readable \emph{between-config} variance (different mounts/backgrounds), which any encoder captures before decoding any pose. (ii)~The POC numbers are noisy/under-fit ($\pm0.075\!\rightarrow\!\pm0.010$, $7\times$ tighter). (iii)~PCA 256 improved the most ($+0.284$), diagnostic of the new between-config variance landing in its top components. (iv)~$z_v$ rose the least ($+0.102$), since it was already near saturation, which is why the margin over raw narrows while the margin over compression stays large. The comparison carrying the cross-modal claim is $z_v$ vs.\ the compression control, decisive in both runs.

\clearpage

\section{Secondary ablations}
\label{sec:moreabl}
Two further ablations, run on the ur5 specialist configuration (5-seed), included for completeness.

\paragraph{Bottleneck size and joint-SIGReg.} On ur5 the latent width has an optimum, where $d{=}128$ gives $0.356/0.147$ (motor/ee), $d{=}256$ gives $0.324/0.156$ (the ur5 specialist of Table~\ref{tab:phase1matrix}), and $d{=}512$ degrades to $0.255/0.122$ with unstable RankMe (some seeds collapse), so $d{\in}[128,256]$ is preferred over a wider bottleneck. Removing the joint-SIGReg term slightly improves the ur5 probes ($0.357/0.173$ vs.\ $0.324/0.156$ with it), consistent with its role as insurance rather than a learning signal and with the healthy RankMe band (Section~\ref{sec:phase1}) showing $z$ does not collapse at this scale; as noted in Section~\ref{sec:objective} we keep the term against collapse at wider bottlenecks, where the $d{=}512$ instability above shows the pressure begins to matter.

\paragraph{Latent geometry (triplet accuracy and alignment/uniformity).} A triplet probe (positive $=$ same scene, nearby tick; negatives laddered from temporally-distant same-scene to different-task) saturates for every feature at the cache's $\sim$1.5--3~s chunk spacing. Kepler-Encoder-v0.1 scores $0.95$--$0.99$ across tiers (5 seeds) and every frozen vision baseline scores $0.98$--$1.00$, since pixel-faithful features pass temporal-metadata tiers trivially, so the probe does not separate abstraction quality and the discriminating geometry measurement is the distance correlation of Section~\ref{sec:mainresults}. Alignment and uniformity~\citep{wangisola2020} describe the trade behind this. PT-ViT places temporal neighbors nearly on top of each other (alignment $0.007$) because all of its embeddings are similar (uniformity $-0.68$), while the fused latent pays a higher alignment ($0.231$) for far better spread ($-2.04$). A probe-based state-retrieval test (rank held-out candidates by predicted end-effector state) places the finetuned head last on every embodiment but mixes PT-ViT and the fused latent at the top with low absolute recall (R@10 $\le 0.09$ against pools of 10--27k), so we do not use it as a headline metric.

\clearpage

\section{Continuous-time, multi-rate extension (planned)}
\label{sec:temporal}
\begin{figure}[!htbp]
  \centering
  \includegraphics[width=0.9\linewidth]{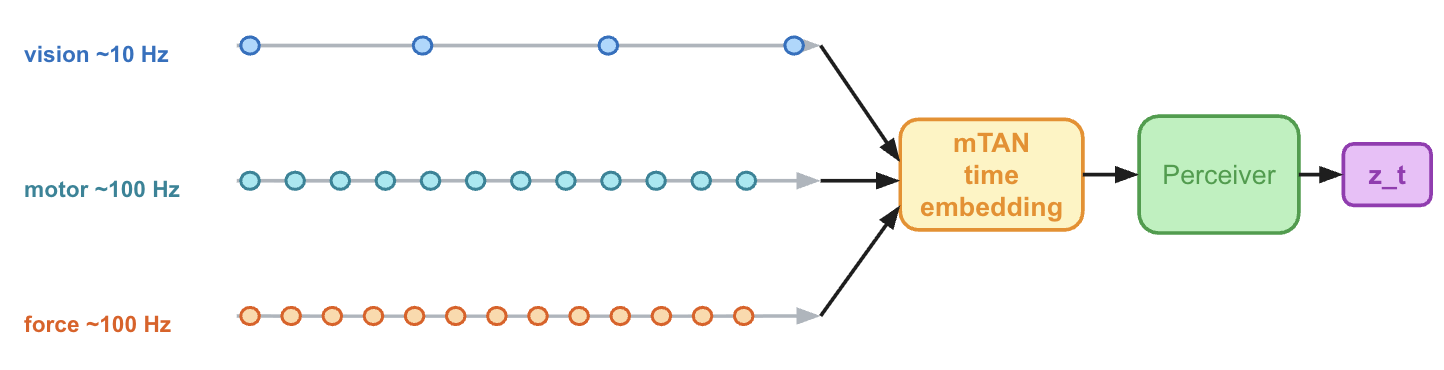}
  \caption{The continuous-time, multi-rate extension. Streams arrive at different native rates (vision $\sim$10~Hz, proprioception and force $\sim$100~Hz), each sample carrying an mTAN continuous-time embedding of its real timestamp, so the Perceiver fuses them by real time with no resampling to a common clock and no zero-padding. The single-timestep model of this paper is the one-tick special case.}
  \label{fig:temporal}
\end{figure}

The single-timestep model of this report validates modality fusion; the central architectural hypothesis, left to future work, is fusing streams \emph{across time at their native rates}. The data side already supports it (tick-anchored native-rate windows with per-token timestamps, Section~\ref{sec:data}); the remaining work is model-side (Figure~\ref{fig:temporal}). Beyond the motor and force payoff expected from time-derivatives, the read-out transfer failure of Table~\ref{tab:transfer} gives this extension a second concrete target, restoring read-out transfer under transient sensor dropout (Section~\ref{sec:discussion}).

The continuous-time embedding encodes each token's timestamp. We embed each token's real timestamp with a Fourier / Time2Vec feature map~\citep{time2vec2019} using fixed, \emph{log-spaced} frequencies, upgradable to learned frequencies (mTAN proper~\citep{mtan2021}) behind the same interface. We deliberately take only mTAN's continuous-time \emph{embedding} and not the whole network, because the Perceiver already supplies mTAN's other half. The learned queries are the reference points and cross-attention is attention-to-observations, a point COPER makes architecturally explicit~\citep{coper2022}. The decisive knob is the frequency band, which must span $\sim$ms (100~Hz force) to seconds (multi-second episodes).

Tokenization varies by stream density. Dense force/torque ($\sim$100--125~Hz) is tokenized by a 1-D CNN over time, which is locally near-regular within a chunk, with the validity mask carried as extra channels so that variable-DOF masking is preserved. Sparse streams such as vision and joints at $\sim$10~Hz remain native tokens with continuous-time embeddings. Context windows are selected by $\Delta t$ from cached timestamps, and masking spans (modality $\times$ time) to predict held-out-time or future ee latents. Causality is a property of the attention mask, causal for the prediction head and any streaming deployment, with causal-versus-bidirectional tested for the representation objective, rather than something baked into the tokenizer.

\paragraph{SIGReg under time.} One caveat is specific to this extension. SIGReg's Epps--Pulley statistic is a \emph{distributional} test over a pool of samples that pushes the \emph{marginal} embedding distribution toward isotropy, and applied to a time-pooled latent its isotropy constraint could wash out structured temporal variance. This does not threaten the design, because SIGReg is an anti-collapse \emph{regularizer}, not the learning signal; temporal structure is carried by the prediction objective (masking over modality~$\times$~time, future-latent prediction) and the recurrent carry-forward, not by the regularizer. To keep the statistic from pooling away time we apply SIGReg to the \emph{per-timestep} latent (each instant's marginal), never to a time-pooled one, and treat its placement and strength as an on/off ablation.

A decisive ablation follows from this design. We compare continuous-time embedding with no resampling against resample-everything-then-1D-CNN, the classical DSP approach. This test validates or rejects the central claim, and we build both and adopt the simpler one if it wins.

\end{document}